%% file: SMPI.tex
\documentclass{article}

\usepackage{natbib}
\usepackage[english]{babel}

\usepackage[letterpaper,top=2cm,bottom=2cm,left=3cm,right=3cm,marginparwidth=1.75cm]{geometry}

\usepackage{amsmath}
\usepackage{graphicx}

\usepackage{setspace}

\usepackage{authblk}
\usepackage[colorlinks=true, allcolors=blue]{hyperref}

\usepackage{physics}
\usepackage[table, dvipsname]{xcolor}         
\usepackage[linesnumbered,ruled,vlined]{algorithm2e}
\usepackage{algorithmic}
\usepackage{caption}
\usepackage{subcaption}
\usepackage{tikz}
\usetikzlibrary{shapes.misc}
\usetikzlibrary{calc}
\usetikzlibrary{arrows}
\usetikzlibrary{shapes}
\usetikzlibrary{decorations.pathreplacing,decorations.markings,decorations.text}

\definecolor{reddark}{RGB}{179,0,59}
\definecolor{greendark}{RGB}{0,225,0}
\definecolor{greendark2}{RGB}{0,200,0}
\definecolor{purple2}{RGB}{147, 81, 22}
\definecolor{grey2}{RGB}{144, 148, 151}
\definecolor{yellow2}{RGB}{255, 216, 1}

\usepackage{fp}
\usepackage{xfp}

\input{math_commands.tex}

\title{Selective Multiple Power Iteration: from Tensor PCA to gradient-based exploration of landscapes}

 \author[1]{Mohamed Ouerfelli\thanks{Mohamed-oumar.ouerfelli@cea.fr}}
    \author[2]{Mohamed Tamaazousti\thanks{Mohamed.tamaazousti@cea.fr}}

  \author[3]{Vincent Rivasseau\thanks{Vincent.rivasseau@ijclab.in2p3.fr}}

\affil[1,2]{Université Paris-Saclay, CEA, List, F-91120 Palaiseau, France}
\affil[2]{Université Paris-Saclay, CNRS/IN2P3, IJCLab, 91405 Orsay, France}
\date{}

\begin{document}
\maketitle

\begin{abstract}

We propose Selective Multiple Power Iterations (SMPI), a new algorithm to address the important Tensor PCA problem that consists in recovering a spike $\vv_0^{\otimes k}$ corrupted by a Gaussian noise tensor $\tZ \in (\mathbb{R}^n)^{\otimes k}$ such that $\tT=\sqrt{n} \beta \vv_0^{\otimes k} + \tZ$ where $\beta$ is the signal-to-noise ratio. SMPI consists in generating a polynomial number of random initializations, performing a polynomial number of symmetrized tensor power iterations on each initialization, then selecting the one that maximizes $\langle \tT, \vv^k \rangle$. Various numerical simulations for $k=3$ in the conventionally considered range  $n \leq 1000$, where existent algorithms exhibit negligible finite size effects, show that the experimental performances of SMPI improve drastically upon existent algorithms and becomes comparable to the theoretical optimal recovery. We show that these unexpected performances are due to a powerful mechanism in which the noise plays a key role for the signal recovery and that takes place at low $\beta$. Furthermore, this mechanism results from five essential features of SMPI that distinguish it from previous algorithms based on power iteration. These remarkable results may have strong impact on both practical and theoretical applications of Tensor PCA. (i) We provide in the supplementary material multiple variants of this algorithm to tackle low-rank CP tensor decomposition. These proposed algorithms also outperforms existent methods even on real data which shows a huge potential impact for practical applications. (ii) We present new theoretical insights on the behavior of SMPI and gradient descent methods for the optimization in high-dimensional non-convex landscapes that are present in various machine learning problems. (iii) We expect that these results may help the discussion concerning the existence of the conjectured statistical-algorithmic gap.

\end{abstract}

\section{Introduction}

Tensor PCA was introduced in the pioneer work of \cite{richard2014statistical} and consists in recovering a signal spike $\vv_0^{\otimes k}$ that has been corrupted by a noise tensor $\tZ$: $\tT =\tZ + \beta \vv_0 ^{\otimes k}$ where $\vv_0$ is a unitary vector and $\beta$ the Signal-to-Noise Ratio (SNR).

The motivation for Tensor PCA is three-fold: 

1) Tensor PCA could be considered as a simple case of Tensor decomposition. However, it has a different motivation which is the theoretical study of the computational limitations in the very low SNR regime while the Tensor decomposition literature mainly address practical applications, often in a large SNR. Yet, algorithms developed for Tensor PCA could be generalised to address Tensor decomposition as in \cite{wang2016online}. Tensor decomposition has various applications such as topic modelling \citep{anandkumar2015learning}, community detection \citep{anandkumar2013tensor},  etc. CP Tensor decomposition also proved useful in the context of deep neural networks, particularly in compressing Convolutional Neural Networks to reduce the memory and the computational cost \cite{astrid2017cp, wang2020cpac}. 
    
    
2) In addition to that, Tensor PCA is also often used as a prototypical inference problem for the theoretical study of the computational hardness of optimization in high-dimensional non-convex landscapes, in particular using the well spread gradient descent algorithm and its variants (\cite{arous2020algorithmic, mannelli2019passed,mannelli2019afraid,mannelli2020marvels}). Indeed, these algorithms are used with great empirical success in many ML areas such as Deep Learning, but unfortunately they are generally devoid of theoretical guarantees. Understanding the dynamics of gradient descent methods in specific landscapes such as Tensor PCA could bring new insights.
    
3) One of the main characteristic of Tensor PCA is its conjectured statistical algorithmic gap: while information theory shows that is theoretically possible to recover the signal for $\beta \sim O(1)$, all existent algorithms have been shown or conjectured to have an algorithmic threshold for $k\geq 3$ of at least $\beta \sim O(n^{(k-2)/4})$. Thus Tensor PCA is considered as an interesting study case of such a gap that appears in various other problems (see references in \citep{arous2020algorithmic} and \citep{luo2020open}).


\paragraph{Contributions:} 
In this paper, (i) we exhibit a novel and powerful convergence mechanism that takes place in small SNR where the noise plays a crucial role in the recovery of the signal. (ii) we provide necessary conditions for this mechanism to be triggered (iii) we propose a new algorithm that we call SMPI for Selective Multiple Power Iterations: that is based on this mechanism. (iv) We show and discuss the potential impact of this algorithm on the three main motivations of Tensor PCA.

\paragraph{Notations}
We use bold characters $\displaystyle \tT, \mM, \vv$ for tensors, matrices and vectors and $\etT_{ijk}, \emM_{ij}, \evv_i$ for their components. $[p]$ denotes the set $\{1,\dots,n\}$. A real $p-$th order tensor $ \tT \in \bigotimes_{i=1}^p \mathbb{R}^{n_i}$ is a member of the tensor product of Euclidean spaces $\mathbb{R}^{n_i}, i \in [p]$. It is symmetric if $\etT_{i_1 \dots i_k}=\etT_{\tau(i_1) \dots \tau(i_k)} \; \forall \tau \in \mathcal{S}_k$ where $\mathcal{S}_k$ is the symmetric group of degree $k$. For a vector $\mathbf{v} \in \mathbb{R}^n$, $\vv^{\otimes p} \equiv \vv \otimes \mathbf{v} \otimes \dots \otimes \vv \in \bigotimes^p \mathbb{R}^n$ denotes its $p$-th tensor power.

We denote $ \tT  (\vv, \vw, \vz) \equiv \langle \tT,\vv \otimes \vw \otimes \vz \rangle \equiv \sum_{ijk} \etT_{ijk} {\evv}_i {\evw}_j {\evz}_k $ the euclidean product between the two tensors $\tT$ and $\vv \otimes \vw \otimes \vz$. Similarly $\tT  (:, \vw, \vz)$ is the vector whose $i$-th entry is $\sum_{jk} \etT_{ijk} {\evw}_j {\evz}_k$ where $e_i$ is the $i$-th basis vector and $\tT  (:, :, \vz)$ is the matrix whose $i,j$th element is $\sum_{k} \etT_{ijk} {\evz}_k$.



\paragraph{Simple and symmetrized power iteration}

The simple power iteration consists in performing the following operation:
\begin{equation}
   \vv \leftarrow \frac{\tT(:,\vv,\vv)}{\norm{\tT(:,\vv,\vv)}}
\end{equation}
Given a non-symmetrical tensor, we define the symmetrized power iteration as:
\begin{equation}
   \vv \leftarrow \frac{\tT(:,\vv,\vv) + \tT(\vv,:,\vv) + \tT(\vv,\vv,:)}{\norm{\tT(:,\vv,\vv) + \tT(\vv,:,\vv) + \tT(\vv,\vv,:)}}
\end{equation}
It is straightforward to see that performing a symmetrized power iteration on a tensor $\tT$ amounts to perform a simple power iteration on the symmetrized tensor  ${\tT_{\text{sym}}}_{i_1 i_2 \dots i_k} \equiv \sum_{\sigma \in \mathcal{S}_k} \tT_{\sigma(i_1) \sigma(i_2) \dots \sigma(i_k)} $.
\begin{equation}
   \vv \leftarrow \frac{\tT_{\text{sym}}(:,\vv,\vv)}{\norm{\tT_{\text{sym}}(:,\vv,\vv)}}
\end{equation}

Unless specified otherwise, we restrict ourselves to a symmetrical tensor with $k=3$, that can easily be generalized to an asymmetrical tensor (by symmetrizing the tensor) and to $k\geq 4$.


\section{Related work}


\paragraph{Tensor PCA}
Sevelar algorithms have been developed to tackle the tensor PCA problem. \cite{richard2014statistical} analyzed many algorithms theoretically and empirically (in a range of $25 \leq n \leq 800$). The tensor unfolding algorithm showed an empirical threshold of $\beta \sim n^{1/4}$ while naive power iteration with a random initialization performed much worse with an empirical threshold of $n^{1/2}$. \cite{hopkins2016fast} provided an algorithm based on sum-of-squares, which was the first with theoretical guarantees whose threshold matches $n^{1/4}$. Other studied methods have been inspired by different perspectives like homotopy in \cite{homotopy17a}, statistical physics (\cite{arous2020algorithmic}, \cite{wein2019kikuchi} and \cite{biroli2020iron}), quantum computing \cite{Hastings2020classicalquantum}, low-degree polynomials \cite{kunisky2019notes}, statistical query \cite{dudeja2021statistical}, random tensor theory  \cite{Ouerfelli2022random} as well as renormalization group \cite{lahoche2021field}.

\paragraph{Power iteration for Tensor PCA}
Power iteration is a simple method that has been extensively used in multiple tensor problems \cite{anandkumar2012tensor,anandkumar2013tensor}. \cite{richard2014statistical} investigated the empirical performance of power iteration with a random initialization in the range of $n \in [50,800]$ and observed an empirical threshold of $n^{1/2}$. Through an improved noise analysis, \cite{wang2016online} showed that for a symmetrical tensor, power iteration is indeed able to recover the signal for a SNR $\beta$ above $n^{1/2}$ with a constant number of initialization and a number of iterations logarithmic on $n$. Their experiments in the range of $n\in [25,\dots,250]$ suggested that this threshold is tight. A recent paper \cite{huang2020power} investigates the simple power iteration for a non-symmetric tensor for Tensor PCA, they prove that the algorithmic threshold is strictly equal to $n^{1/2}$. The results of their experiments for $n \in [200,\dots,800 ]$ matches their theoretical results. In this paper, we aim to draw attention to a surprising observation that contrasts with previous work: if we impose five essential features for an algorithm based on power iteration or gradient descent (use a symmetrized power iteration, impose a polynomial number of initializations and iterations, etc.), we observe that a novel powerful mechanism for the convergence towards the signal takes place, leading to a fundamentally different performance. In fact, for $n \in [50,1000]$, SMPI is the first algorithm to exhibit an empirical threshold corresponding to $O(1)$ and whose results matches the theoretically-optimal correlation at large $n$. 

\paragraph{Prototypical inference problem and exploration of complex landscapes}
Tensor PCA, as well as a weaker version of it, the matrix-tensor PCA introduced in \cite{mannelli2020marvels}, has been considered by the scientific community as a prototypical inference problem in order to analyze the interplay between the loss landscape and performance of descent algorithms \citep{mannelli2019passed,mannelli2020marvels,arous2020algorithmic}. However, these studies focused on the gradient flow and Langevin dynamics. We focus here on power iteration which allows us to uncover the importance of a large step size in order to avoid spurious minima and to converge towards the signal. Note that \cite{choromanska2015loss} investigated a very similar landscape to Tensor PCA in order to gain insights on neural network landscapes. 
\paragraph{The conjectured statistical algorithmic gap: computational hardness}
While \cite{richard2014statistical} proved that the optimal theoretical threshold is of order $\beta_\text{opt} = O(1)$, many of the suggested methods have been predicted (based on empirical results for $25 \leq n \leq 1000$) to have at best an algorithmic threshold of $O(n^{1/4})$. This led to a conjecture that it is not possible to achieve the recovery of the signal vector with polynomial time for a $\beta$ below $O(n^{1/4})$. A rich theoretical literature emerged in order to understand the fundamental reason behind the apparent computational hardness of Tensor PCA. Average-case reduction has been investigated in \cite{brennan2020reducibility,luo2020open}. While several papers (such as \cite{perry2020statistical,lesieur2017statistical,ros2019complex,jagannath2020statistical}) provided new results on the statistical threshold of Tensor PCA, there have been many results for thresholds of specific algorithmic models. In particular, \cite{kunisky2019notes} proves the failure of low-degree methods for $1 \ll \beta \ll n^{1/4}$ and surveys a recent and interesting line of research that explores the conjecture that the failure of low-degree methods indicates the existence of statistical algorithmic gap in high-dimensional
inference problems. \cite{arous2020algorithmic} provides a possible explanation for the failure of the Langevin dynamics and gradient descent (in the infinitely small learning rate limit) which is another class of algorithms.

\section{Selective Multiple Power Iteration (SMPI)}

\subsection{General Principle of SMPI}


The proposed SMPI (Algorithm 1) consists in applying, in parallel, the power iteration method with $m_\text{iter}$ iterations to $m_\text{init}$ different random initialization. Then, SMPI uses the maximum likelihood estimator to select the output vector in this subset by choosing the vector that maximizes $\tT ( \vv,\vv,\vv)$.

\begin{algorithm}[h!]
   \caption{Selective Multiple Power Iteration}
    \label{algo:recovery}
\begin{algorithmic}[1]\onehalfspacing
   \STATE {\bfseries Input:} The tensor $\tT=\tZ + \beta \vv_0^{\otimes k}$, $m_\text{init}>10n$, $m_\text{iter}>10n$,$\Lambda$
   \STATE {\bfseries Goal:} Estimate $\vv_0$.
   
   \FOR{ i=0 to $m_\text{init}$} 
   \STATE Generate a random vector $\vv_{i,0}$
   \FOR{ j=0 to $m_\text{iter}$} 
   \STATE $\displaystyle\vv_{i,j+1}=\frac{\tT(:,\vv_{i,j},\vv_{i,j})}{\norm{\tT(:,\vv_{i,j},\vv_{i,j}) }}$
   \IF{ $\displaystyle j>\Lambda$ and $ \abs{\langle \vv_{i,j-\Lambda},\vv_{i,j}\rangle}\geq 1-\varepsilon $}
   \STATE $\vv_{i,m_\text{iter}}=\vv_{i,j}$
   \STATE \textbf{break}
\ENDIF
   \ENDFOR
\ENDFOR
   \STATE Select the vector $\vv=\arg \max_{1\leq i \leq {m_\text{init}}} \tT (\vv_{i,m_\text{iter}} ,\vv_{i,m_\text{iter}} ,\vv_{i,m_\text{iter}})$ 
   \STATE {\bfseries Output:} the estimated vector $\vv$
\end{algorithmic}
\end{algorithm}

\paragraph{Link between Power Iteration and Gradient descent}
Maximising $\tT (\vv, \vv, \vv)$ is equivalent to finding the minimum of the cost function defined as $H(\vv)\equiv -\tT (\vv, \vv, \vv)$ \cite{biroli2020iron}.

If we consider the function $H(\vv)=-\tT (\vv, \vv, \vv)$ defined on $\vv \in \mathbb{R}^n$, its gradient on $\mathbb{R}^n$ is $\nabla H = 3 \tT (:,\vv ,\vv)$. However, given that we restrict ourselves to the manifold $S^1=\{\vv \in \mathbb{R}^n \mid \vv / \norm{\vv}=1 \}$, the gradient has to be projected on the tangent space at $\vv$ (which is the subspace orthogonal to $\vv$) \citep{ros2019complex}. Thus, the gradient is equal to $\vg \equiv \nabla H (\vv) -(\nabla H (\vv).\vv)\vv$ which will be for our model $\vg=-\tT (:,\vv ,\vv) +\tT (\vv,\vv ,\vv).\vv$.
Therefore, power iteration could be written as:
\begin{equation}
\begin{split}
    \displaystyle \vv \leftarrow \frac{\tT (:,\vv ,\vv)}{\norm{\tT (:,\vv ,\vv)}}&=\frac{\tT (:,\vv ,\vv) - \tT (\vv,\vv ,\vv).\vv+\tT (\vv,\vv ,\vv).\vv}{\norm{\tT (:,\vv ,\vv)}}
    =\frac{-\vg+\tT (\vv,\vv ,\vv).\vv}{\norm{\tT (:,\vv ,\vv)}}\\
    &=\frac{\tT (\vv,\vv ,\vv)}{\norm{\tT (:,\vv ,\vv)}}.(\vv-\frac{\vg}{\tT (\vv,\vv ,\vv)})
\end{split}
\end{equation}
We can see that the power iteration could be seen as a gradient descent with an adaptive step size equal to $1/(\tT (\vv,\vv ,\vv))$. This step size has the convenient particularity that it is large for a random $\vv$ but becomes small for vectors $\vv$ such that $\tT (\vv,\vv ,\vv)$ is large, for example when we are close to a minimum of $H$.


\subsection{The essential features of SMPI}
We stress here an important and fundamental difference between our algorithm with previous algorithms based on power iteration. In order for this method to succeed in the large noise (low SNR) setting, we need these five features that, for the best of our knowledge, we are the first to impose:
\begin{enumerate}
    \item Using power iteration or a gradient descent with a large enough step size.
    \item In the case of power iteration and non-symmetric tensor, using the symmetrized version (or equivalently, symmetrize the tensor).
    \item Prohibiting a stopping criteria on two consecutive iterations (such as $1-\abs{\langle \vv_{i-1},\vv_i \rangle}< \varepsilon$ for a given small $\varepsilon>0$) and instead, use a criteria based on non-consecutive iterations distant by $\Lambda$ $1-\langle \vv_{i-\Lambda},\vv_i \rangle< \varepsilon$ for $\varepsilon>0$ and $\Lambda = O(n)$.
    \item Using at least a polynomial number of iterations. 
    \item Using at least a polynomial number of initializations.
\end{enumerate}


\subsection{Experimental results}
In this section, we compare the results of our algorithm with the state of art for every $\beta \in \{1.2,1.3,1.4,1.8,2.2\}$ and for every $n\in \{100,200,400\}$. We averaged over 50 different realizations of $\tT=\tZ + \sqrt{n} \beta \vv^{\otimes 3}$ where $\tZ$ is a tensor with random Gaussian components. We plot in the figure the correlation of the vector $\vv$ output by each algorithm with the signal vector $\vv_0$ and plot the $95\%$ confidence interval bars. The algorithms considered are SMPI (that we perform after symmetrizing the tensor $\tT $), the Homotopy-based algorithm (Hom) \cite{Hom2016arXiv161009322A}, the Unfolding algorithm \cite{richard2014statistical} which are considered as the two main successful algorithms for Tensor PCA, as well as the CP tensor decomposition algorithm of the Python package TensorLy \cite{TensorLy:v20:18-277} used with a rank equal to one.
Similar results are obtained for $n=1000$ (only for SMPI and Homotopy) and on the non symmetric case and are provided in the Appendix.

\begin{figure*}[h!]
\begin{subfigure}{.49\textwidth}
    \centering
    \includegraphics[width=0.95\textwidth]{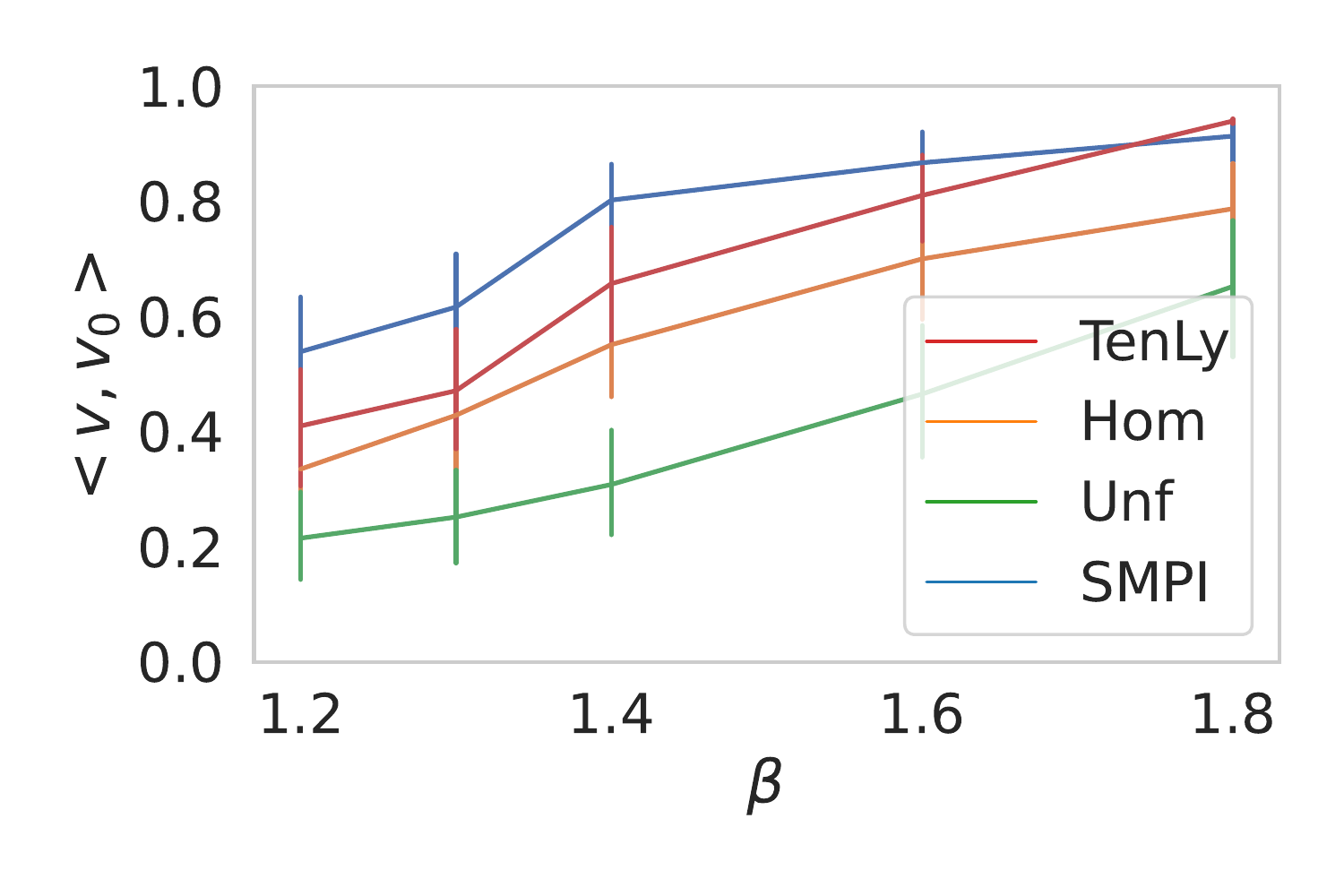}
    \caption{n=50}
    \label{fig:max92}
\end{subfigure}%
\begin{subfigure}{.49\textwidth}
    \centering
    \includegraphics[width=0.95\textwidth]{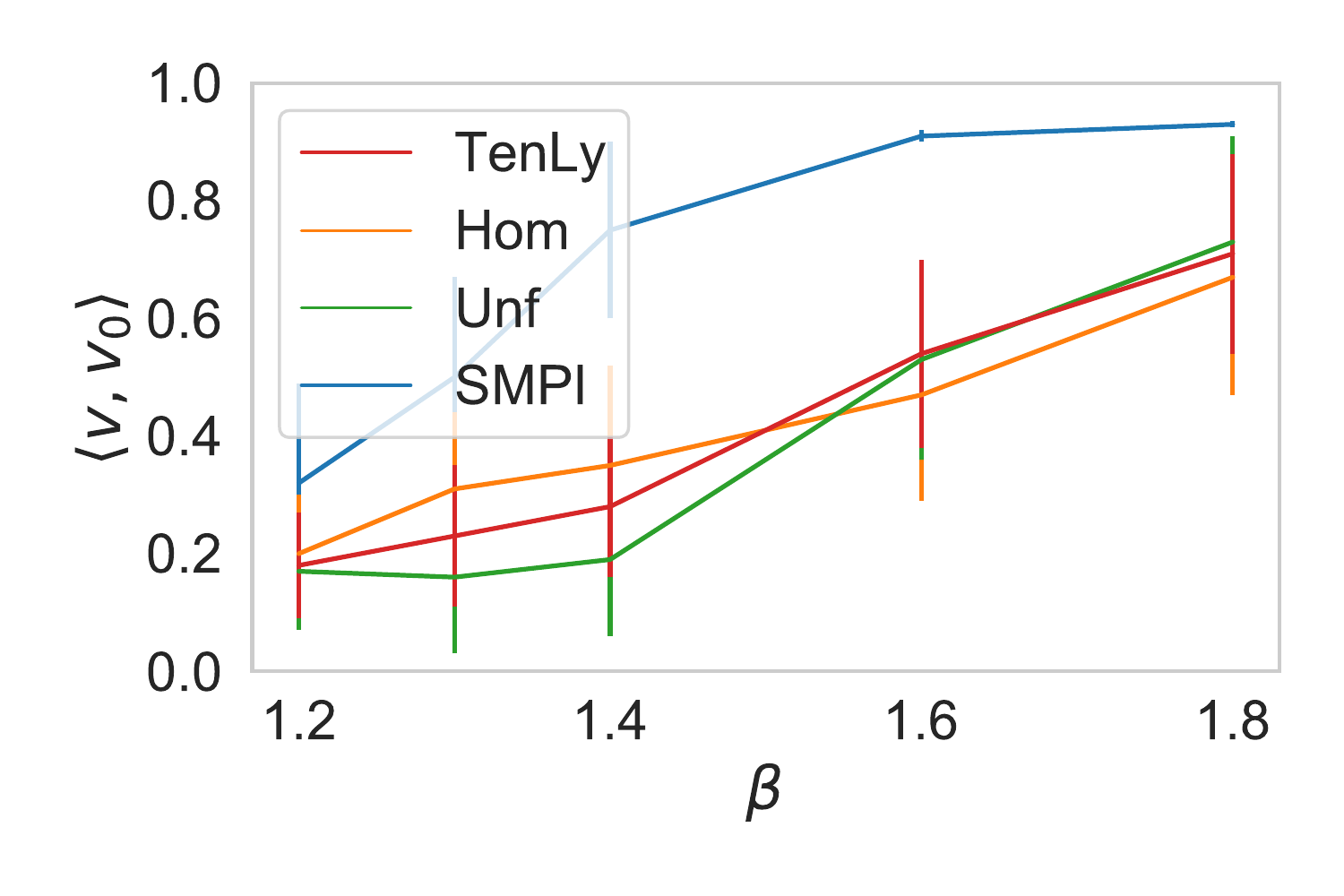}
    \caption{n=100}
    \label{fig:max93}
\end{subfigure}

\begin{subfigure}{.49\textwidth}
    \centering
    \includegraphics[width=0.95\textwidth]{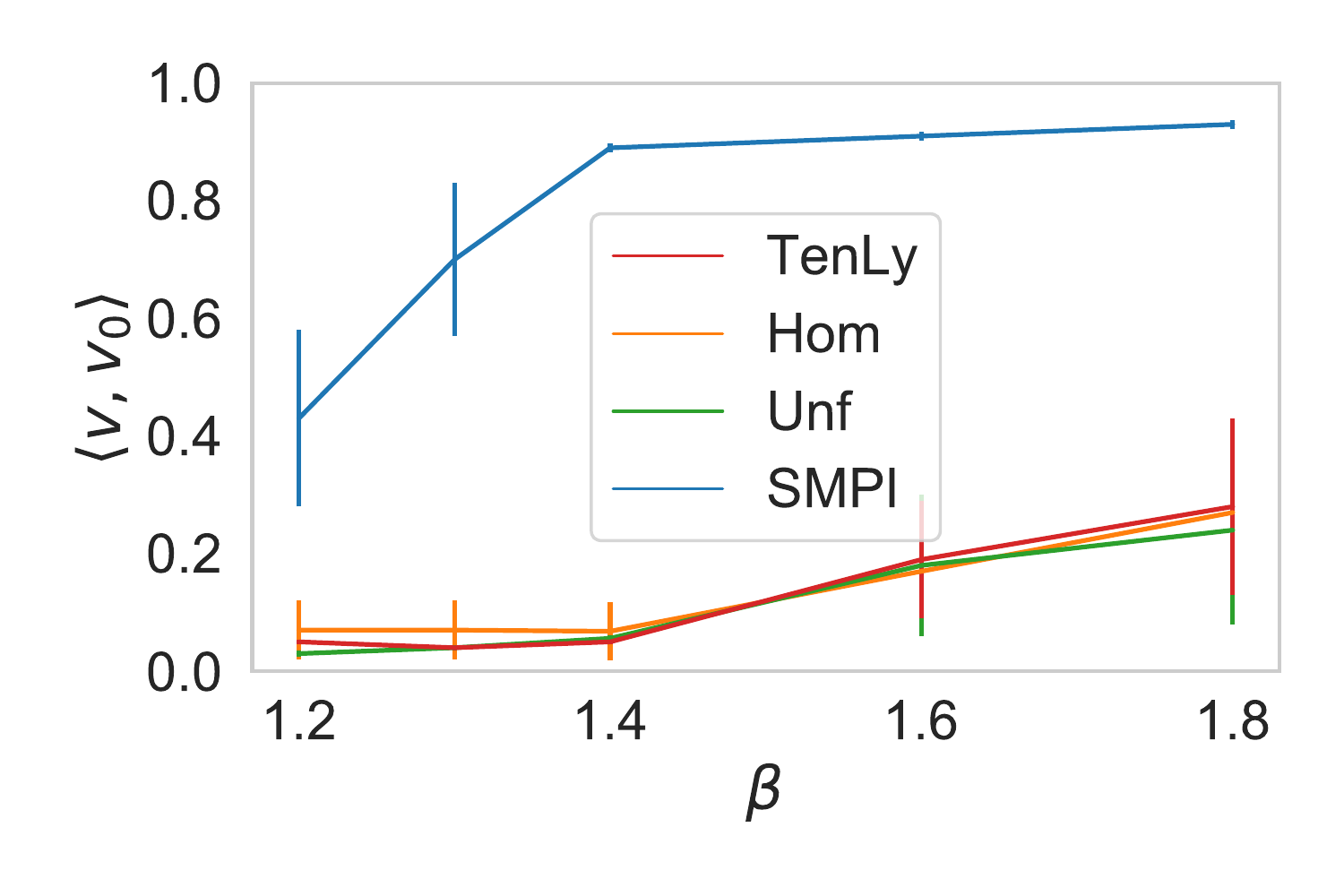}
    \caption{n=200}
    \label{fig:max2}
\end{subfigure}%
\begin{subfigure}{.49\textwidth}
    \centering
    \includegraphics[width=0.95\textwidth]{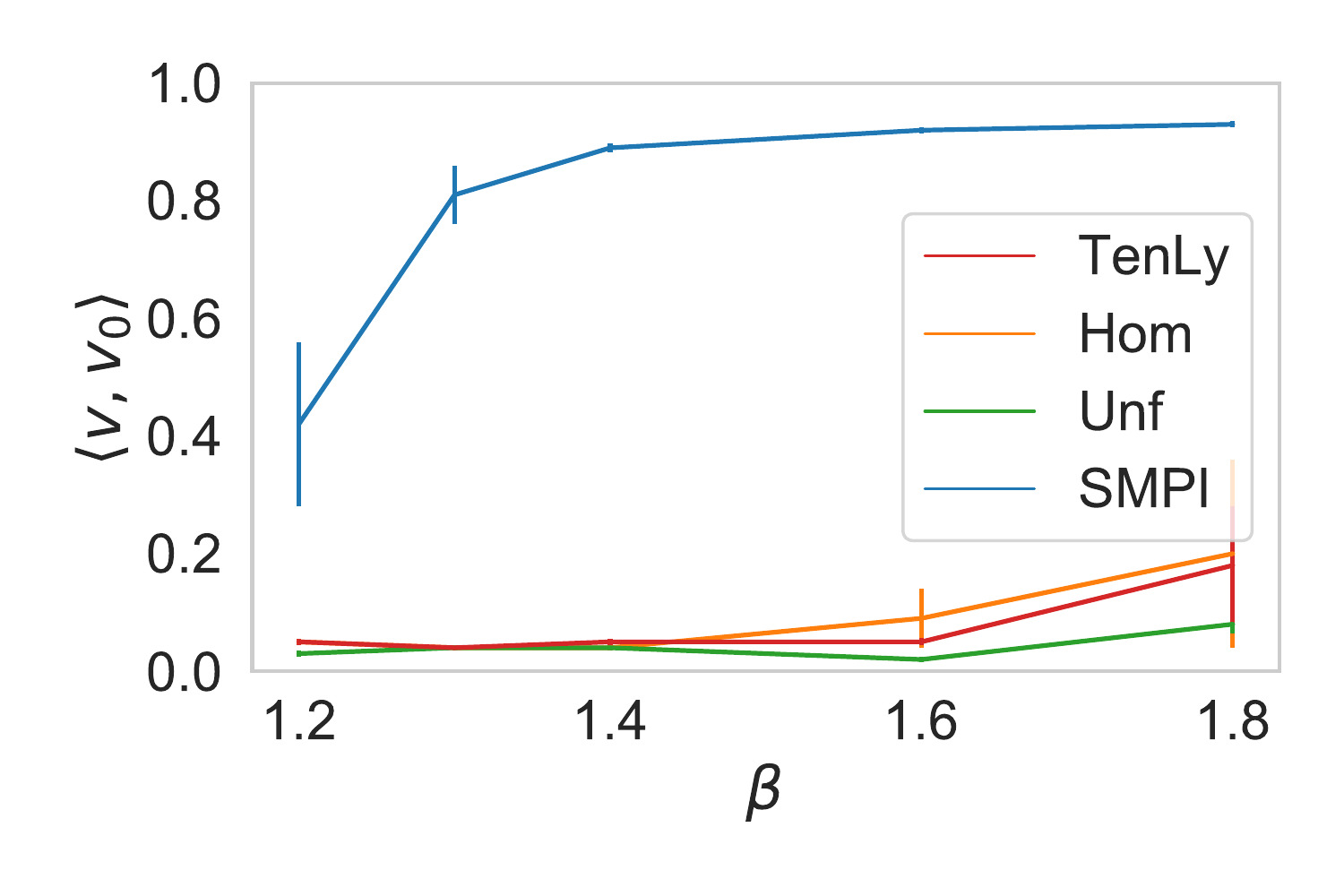}
    \caption{n=400}
    \label{fig:Per2}
\end{subfigure}

\caption{Comparison of the results of SMPI with TensorLy (TenLy) and the State-of-the-art represented here by the Unfolding (Unf) and Homotopy-based (Hom) methods for four values of the dimension of each axe of the tensor ($n=50, 100, 200, 400$). The results consist of the correlation between the output of each algorithm and the signal vector. }
\label{Comparison SOTA}
\end{figure*}

We used a range of $n$ that is commonly considered in empirical investigations of algorithms on Tensor PCA. Indeed, for $n=1000$ the tensor has $10^9$ non-zero entries which becomes extremely costly in memory and computational power. Furthermore, to the best of our knowledge, all algorithms investigated exhibited negligible finite size effectis in this range of $n$ (more details in subsection \ref{Finitesize}). We observe in Figure \ref{Comparison SOTA}, that even for small instances of $n$, like $n=100$, our algorithm performs significantly better than the state of art. The gap between the results of our algorithm and the state of art drastically increases with $n$. 

\paragraph{Complexity of SMPI}

The complexity of SMPI is equal to $ m_\text{init} \times m_\text{iter} \times n^2$. In practice, $m_\text{init}=m_\text{iter}=10n$ already gives us excellent results for a SNR in the range $1<\beta < n^{1/4}$, the complexity is thus $\sim 10 n^4$. It is important to note that such large values of $m_\text{init}$ and $m_\text{iter}$ are necessary as it will be discussed in the following section.


\section{Theoretical insights on the SMPI algorithm}

\subsection{The mechanism: the role of the noise for signal recovery}

In the power iteration, let's denote the part associated to the noise $\vg_N$ and the one associated to the signal $\vg_S$.
\begin{alignat}{3}
    \tT (:,\vv, \vv) &= \tZ (:,\vv, \vv) \;\; &+ \;\;&\beta \langle \vv,\vv_0 \rangle^2 \vv_0\\
        &\equiv \vg_N &+ &\vg_S
\end{alignat}

\begin{figure}[h!]
    \centering
\begin{subfigure}{0.5\textwidth}
    \centering
    \includegraphics[width=\textwidth]{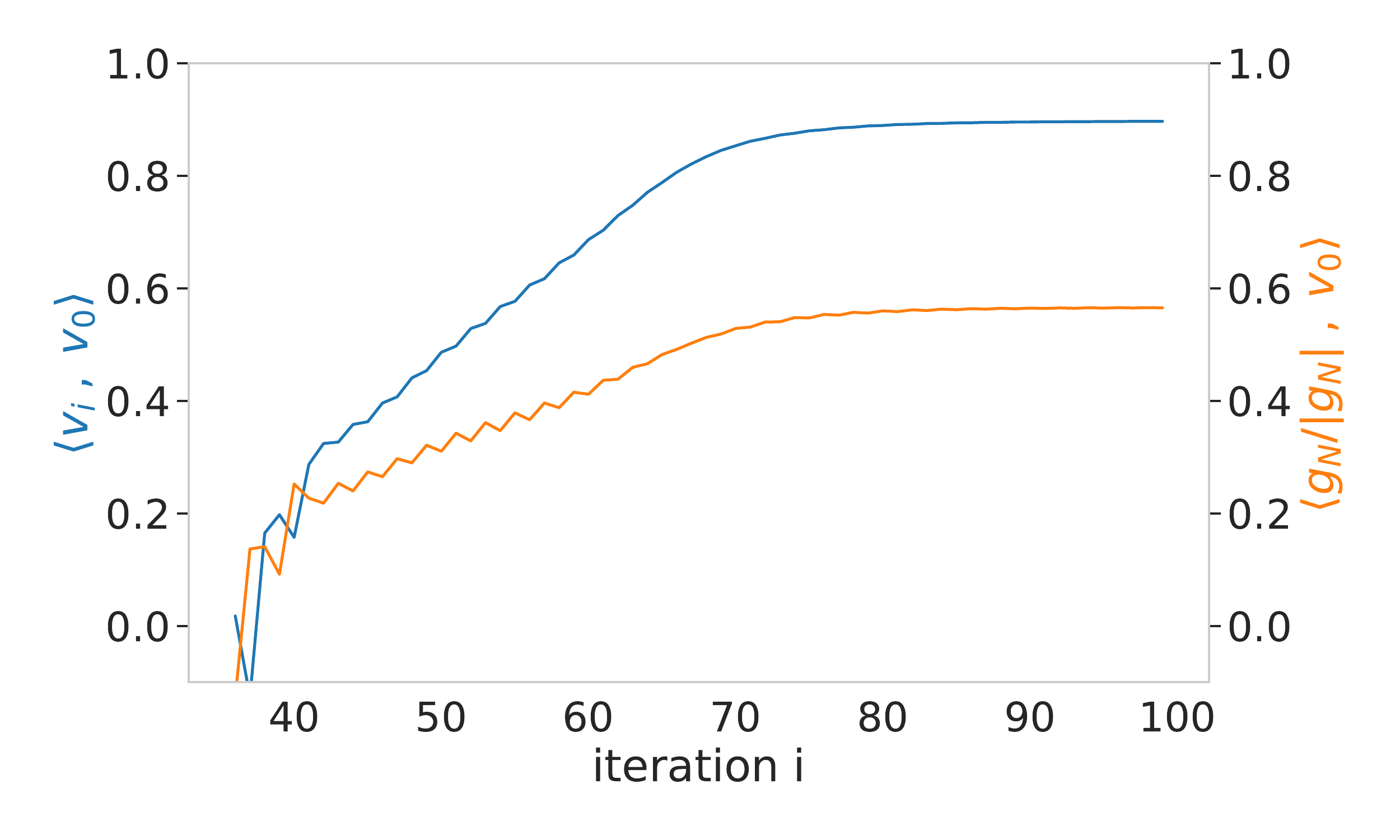}
\end{subfigure}%
\begin{subfigure}{0.5\textwidth}
    \centering
    \includegraphics[width=\textwidth]{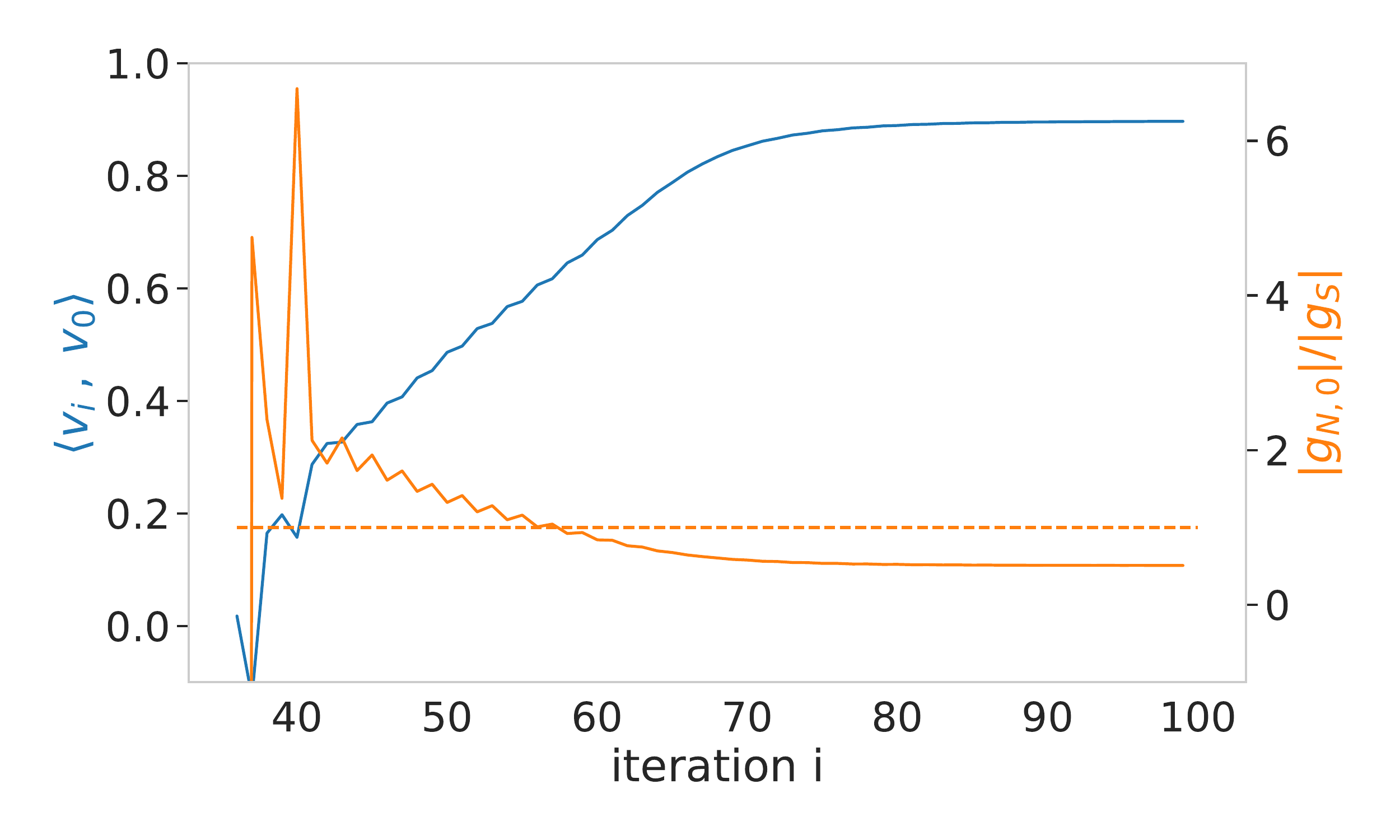}
\end{subfigure}
\caption{In blue, the total correlation of $\vv_i$ with $\vv_0$ at each iteration $i$. In orange in the left, the correlation between the normalized $\vg_N$ and $\vv_0$. In orange in the right, the ratio between the contributions of the noise gradient to $\vv_0$: $\vg_{N,0} \equiv \langle \vg_N, \vv_0 \rangle \vv_0$ and the signal gradient $\vg_S$ that is correlated to $\vv_0$.}
\label{Grad}
\end{figure}

In Figure \ref{Grad}, we illustrate by an example, a pattern observed on all successful convergence in low SNR. For a given random initialization $\vv^1$, we plot in blue (in both the left and the right subfigures) the correlation between the signal $\vv_0$ with $\vv^i$ (obtained after $i$ iterations on $\vv^1$). The horizontal axis begins from the iteration where the convergence towards the signal starts. In the left subfigure, we plot in orange the correlation between $\vv_0$ and $\vg_N/\norm{\vg_N}$, the normalized gradient associated to the noise tensor. In the right we plot in orange the ratio between the contribution of the noise gradient to $\vv_0$: $\vg_{N,0} \equiv \langle \vg_N, \vv_0 \rangle \vv_0$ and the signal gradient $\vg_S$. 

We observe an unexpected result: the gradient $\tZ (:,\vv^i, \vv^i)$ is non-trivially correlated to $\vv_0$ and thus partially converges to $\vv_0$. Moreover, the spikes in the right figure at the beginning of the convergence suggest that it is the term $\tZ (:,\vv^i, \vv^i)$ that triggers the convergence towards $\vv_0$ as it gives the largest contribution to the component of $\vv^i$ correlated to $\vv_0$ at the start of the convergence. This suggests that the gradient associated to the noise actually triggers the convergence towards the signal and carries it all along. 



\paragraph{Towards the quantification of this mechanism}
Let's denote $\vv$ the output of the MLE (i.e. $\vv^*=\arg \max_{\norm{\vv}=1} (\tT(\vv,\vv,\vv)$ ) and $\hat{\vv}$ the output of SMPI. In order to have an indication if the mechanism we illustrated in our simulations is due to finite-size effects or is characteristic of large $n$ values, we can compare the experimental value of the plateau of $\displaystyle \langle \frac{\vg_N}{\norm{\vg_N}},\vv_0 \rangle=\langle\frac{ \tZ (:,\hat{\vv}, \hat{\vv})}{|| \tZ (:,\hat{\vv}, \hat{\vv}) ||},\vv_0 \rangle$ with its theoretical value obtained for $n\rightarrow \infty$ using the following formula (a proof is found in appendix \ref{Appplateau}):
\begin{equation}
    \langle\frac{ \tZ (:,\vv, \vv)}{|| \tZ (:,\vv, \vv) ||},v_0 \rangle =  \frac{\langle v,v_0\rangle(\norm{\tT (:,\vv, \vv)} -  \beta \langle \vv,\vv_0 \rangle)}{\sqrt{\norm{\tT (:,\vv, \vv)}^2 + \beta^2 \langle \vv,\vv_0 \rangle^4-2 \beta \langle \vv,\vv_0 \rangle^3 \norm{\tT (:,\vv, \vv)}}}
    \label{plateau}
\end{equation}

Indeed, \cite{jagannath2020statistical} provided analytic formula for $\norm{\tT (:,\vv^*, \vv^*)}$ and $\langle \vv^*,\vv_0 \rangle$ for a given $\beta$ in the large $n$ limit where $\vv^*$ is the output of the MLE that we use to compare the theoretical expression with the empirical results of the plateau. For $\beta=1.44 \sqrt{n}$, the theoretical value is $\displaystyle\langle \frac{ \tZ(:,\vv^*, \vv^*)}{\norm{\tZ(:,\vv^*, \vv^*)}},v_0 \rangle_\text{th}=0.496$.
The table \ref{tab:plateau} give the average and the standard deviation obtained experimentally for different $n$

\begin{table}[h!]
\small
\setlength{\tabcolsep}{2.3pt}
\caption{Experimental plateau}
\label{tab:plateau}
\begin{center} \begin{tabular}{c|c|c|c|c|c}
\hline
   $n$ &  $50$ &  $100$& $150$&  $200$& $400$ \\
   \hline
   $\displaystyle \langle \frac{ \tZ(:,\hat{\vv}, \hat{\vv})}{\norm{\tZ(:,\hat{\vv}, \hat{\vv})}},\vv_0 \rangle$ &
   \begin{tabular}{c} 
        $0.469 $   \\
        $\pm 0.148$
   \end{tabular}  & 
   \begin{tabular}{c}
        $0.518$   \\
        $\pm 0.074$
   \end{tabular}  &
   \begin{tabular}{c}
        $0.507$   \\
        $\pm 0.056$
   \end{tabular} &
   \begin{tabular}{c}
        $0.487 $   \\
        $\pm 0.038$
   \end{tabular}   &
   \begin{tabular}{c}
        $0.511 $   \\
        $\pm 0.025$
   \end{tabular}   \\
   \hline
   
\end{tabular}
\end{center}
\label{Tab:comp}
\end{table}

We see that the theoretical value $\displaystyle\langle \frac{ \tZ(:,\vv^*, \vv^*)}{\norm{\tZ(:,\vv^*, \vv^*)}} ,\vv_0 \rangle_\text{th}=0.496$ is well inside the error bar and that the standard deviation gets smaller as $n$ grows. This shows an excellent adequacy between the empirical results of the plateau related to $\hat{\vv}$ and the theoretical expectation related to $\vv^*$. For that matter, it suggests that the exploration of the landscape of Tensor PCA by SMPI already reached the large $n$ regime in the experimental values of $n$ considered.

\subsection{Importance of the main features of SMPI}

\subsubsection{Importance of the symmetrized power iteration}
\begin{figure}[h!]
    \centering
\begin{subfigure}{0.45\textwidth}
    \centering
    \includegraphics[width=\textwidth]{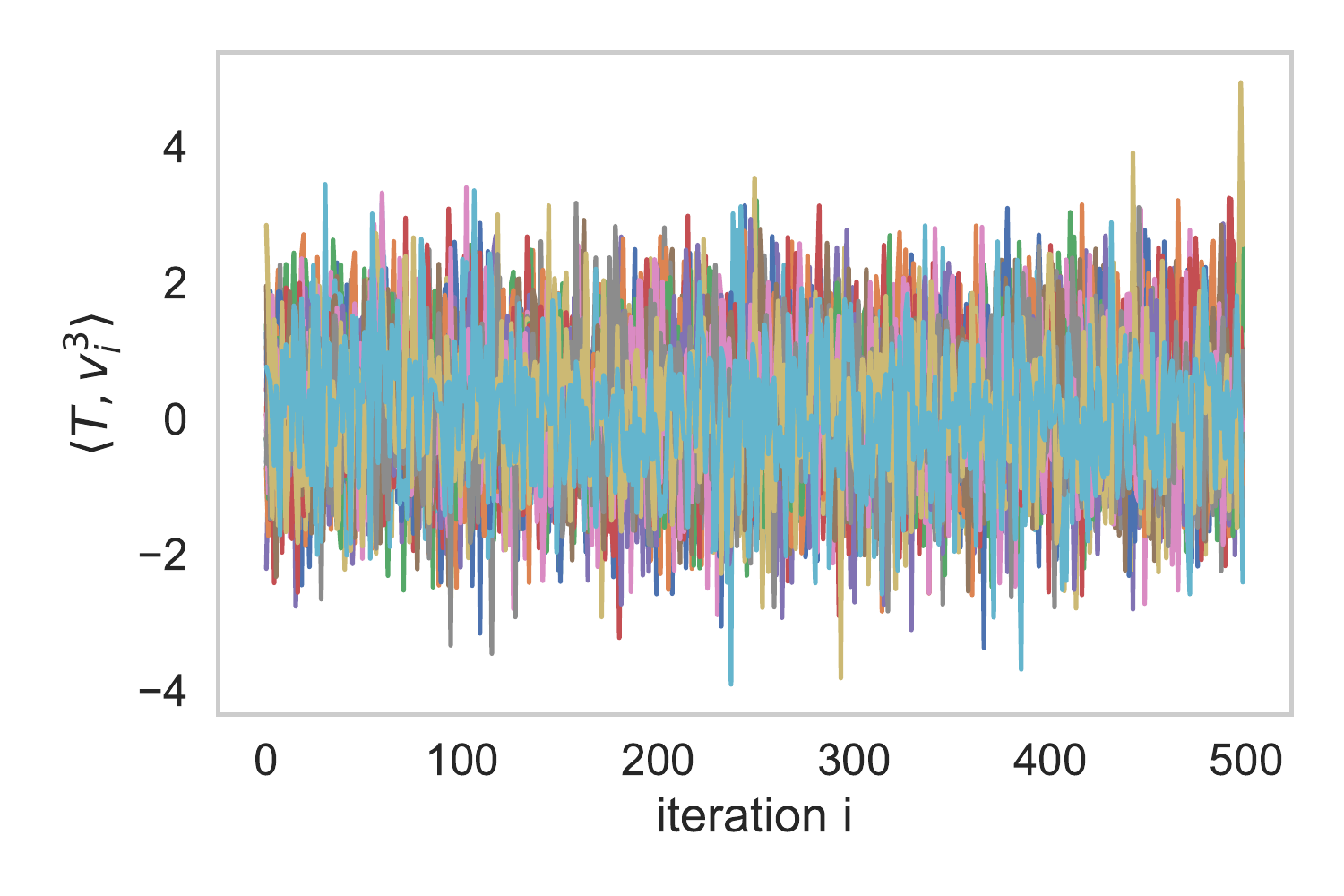}
\end{subfigure}%
\begin{subfigure}{0.45\textwidth}
    \centering
    \includegraphics[width=\textwidth]{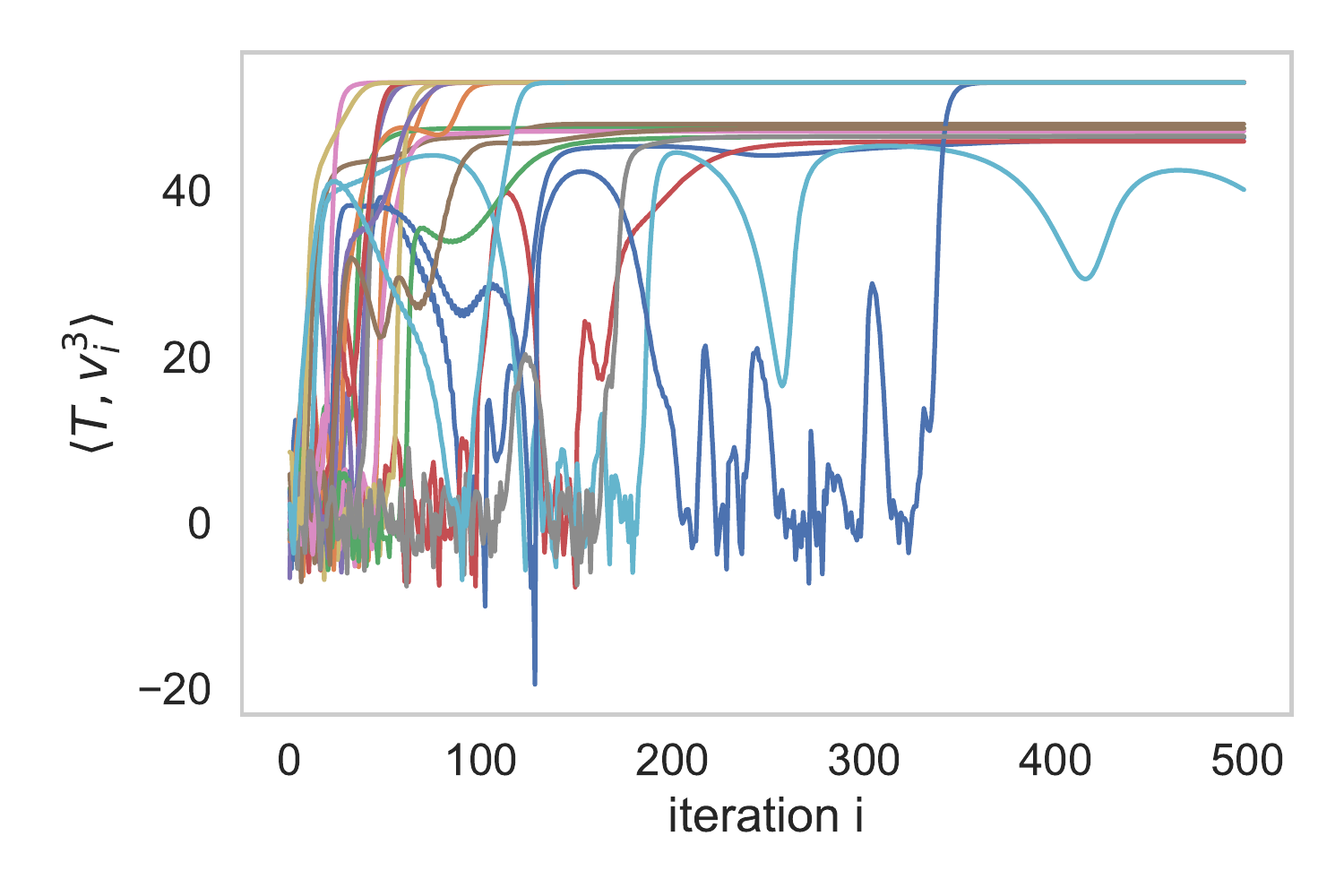}
\end{subfigure}
\caption{For a non-symmetrical tensor $\tT = \tZ + \beta \vv_0^{\otimes 3}$ where $n=200$ and $\beta=1.2 \beta_\text{th}$, we generate $20$ random initializations. For each initialization, we plot with a different color the value of $\tT (\vv_i,\vv_i,\vv_i)$ at each iteration $i$ : in the left using a simple power iteration and in the right using a symmetrized power iteration. We observe a fundamentally different behavior in the case of a symmetrized power iteration.}
\label{fig:SymPI}
\end{figure}

For a non-symmetrical tensor, it has been proven in \cite{huang2020power} that simple power iteration exhibits an algorithmic threshold strictly equal to $n^{1/2}$. For a SNR below this threshold, the output of the power iteration behaves like a random vector at each iteration. In contrast, our numerical experiments suggest that the picture is fundamentally different for a symmetrized power iteration. In Figure \ref{fig:SymPI} we plot $\tT(\vv_i,\vv_i,\vv_i)$ for a simple power iteration (in the left) and the symmetrized power iteration (in the right) for a non-symmetrical tensor and for small SNR ($\beta=1.2 \beta_\text{th}$ where $\beta_\text{th}$ is the theoretical optimal threshold \cite{jagannath2020statistical}). We observe that while the result of simple power iteration matches with the theoretical results obtained in \cite{huang2020power}, the symmetrized power iteration exhibits a fundamentally different behavior that, to the best of our knowledge, has not been fully investigated so far.
    
\subsubsection{The role of a large step size and the stopping condition}

The landscape of the cost function $H(\vv)\equiv -\tT (\vv, \vv, \vv)$ is characterized by an exponentially large number of critical points \cite{ros2019complex}. Although it can't completely explain the success of SMPI, the large step size (or equivalently the use of symmetrized power iteration) is still essential for the gradient descent to escape many of the spurious minima that it may get trapped in. To understand what happens, let's denote $\vm_i$ a minimum of $\vv \rightarrow \tT(\vv,\vv,\vv)$, let's name $\{\vv_i\}_{1\leq i\leq n}$ the eigenvalues of the matrix $\tT (:,:, \vm_i)$. Let's assume that we are in the vicinity of $\vm_i$ and initialize with $\displaystyle \vy_0=\frac{\vm_i+\epsilon \vv_1}{1+ \epsilon^2}$ and let's note $\displaystyle \vy_1=\frac{\tT (:,\vy_0, \vy_0)}{\norm{\tT (:,\vy_0, \vy_0)}}$. 

We can show that if $2\abs{\lambda_1}>\tT(\vm_i,\vm_i,\vm_i)$, the correlation with $\vm_i$ will get smaller after a power iteration at first order of $\varepsilon$:
\begin{equation}
\begin{split}
    \tT(:,\vm_i+\varepsilon \vv_1,\vm_i+\varepsilon \vv_1)&=\tT(:,\vm_i,\vm_i)+2 \varepsilon \tT(:,\vm_i , \vv_1) + O(\epsilon^2)  \\
    &=\tT(\vm_i,\vm_i,\vm_i) \vm_i + \varepsilon \;  2\lambda_1  \vv_1 + O(\epsilon^2)
\end{split}
\end{equation}
Hence, if $2\abs{\lambda_1}>\tT(\vm_i,\vm_i,\vm_i)$ : $\langle \vy_1,\vm_i \rangle < \langle \vy_0,\vm_i \rangle$ and  $\langle \vy_1,\vv_1 \rangle > \langle \vy_0,\vv_1 \rangle$. Which means that if any of the eigenvalues of the Hessian matrix at $\vm_i$ is smaller than $- 2 \tT(\vm_i,\vm_i,\vm_i)$, then the minimum is unstable under power iteration and the algorithm will diverge away from it.

\begin{figure}[h!]
    \centering
    
    \resizebox{0.95\textwidth}{!}{%
\begin{subfigure}{0.54\textwidth}
    \centering
    \includegraphics[width=\textwidth]{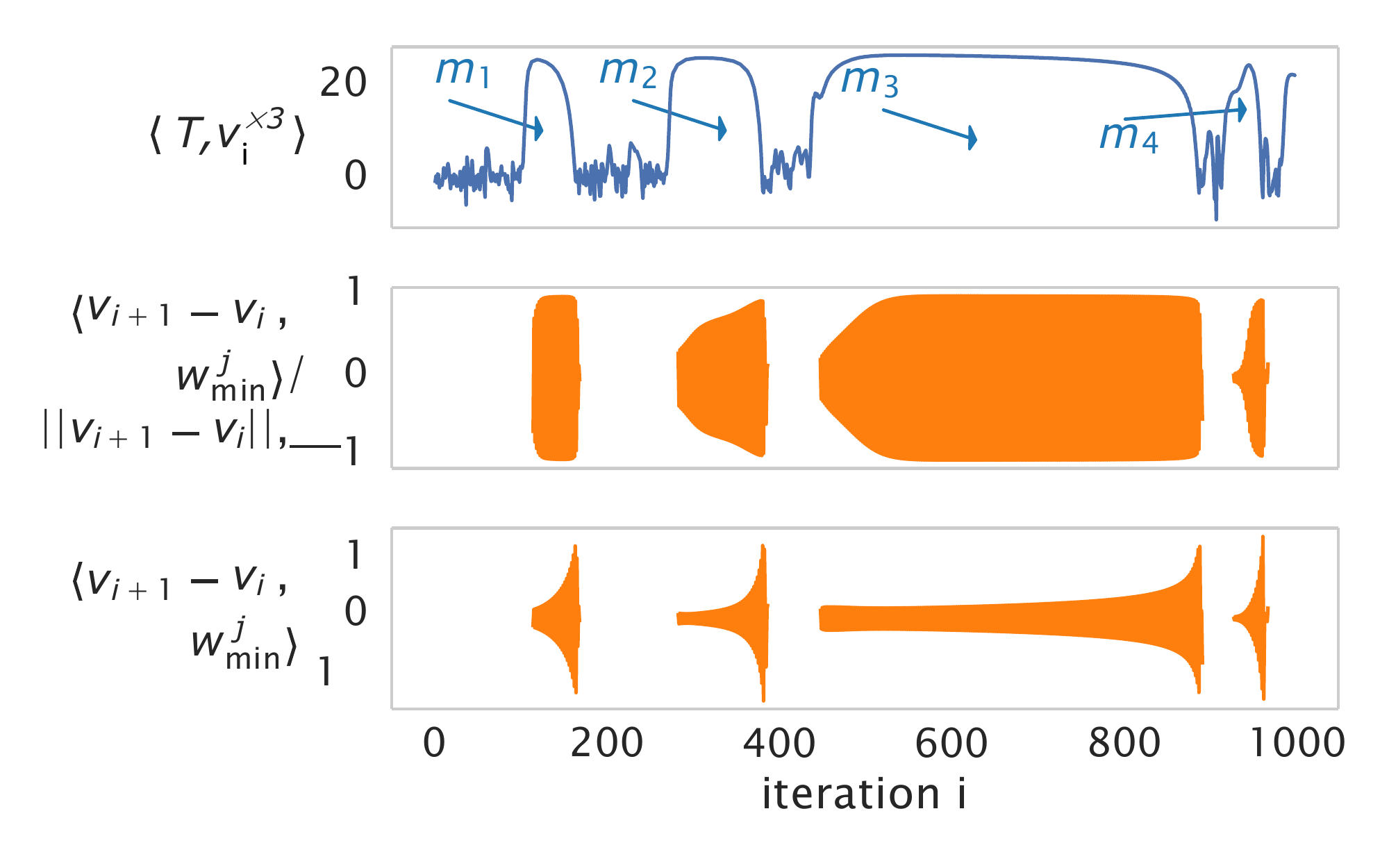}
    \label{fig:Per}
    \caption{}
\end{subfigure}%
\begin{minipage}[H]{0.45\textwidth}

\begin{subfigure}{\textwidth}
    \centering
    \includegraphics[width=\textwidth]{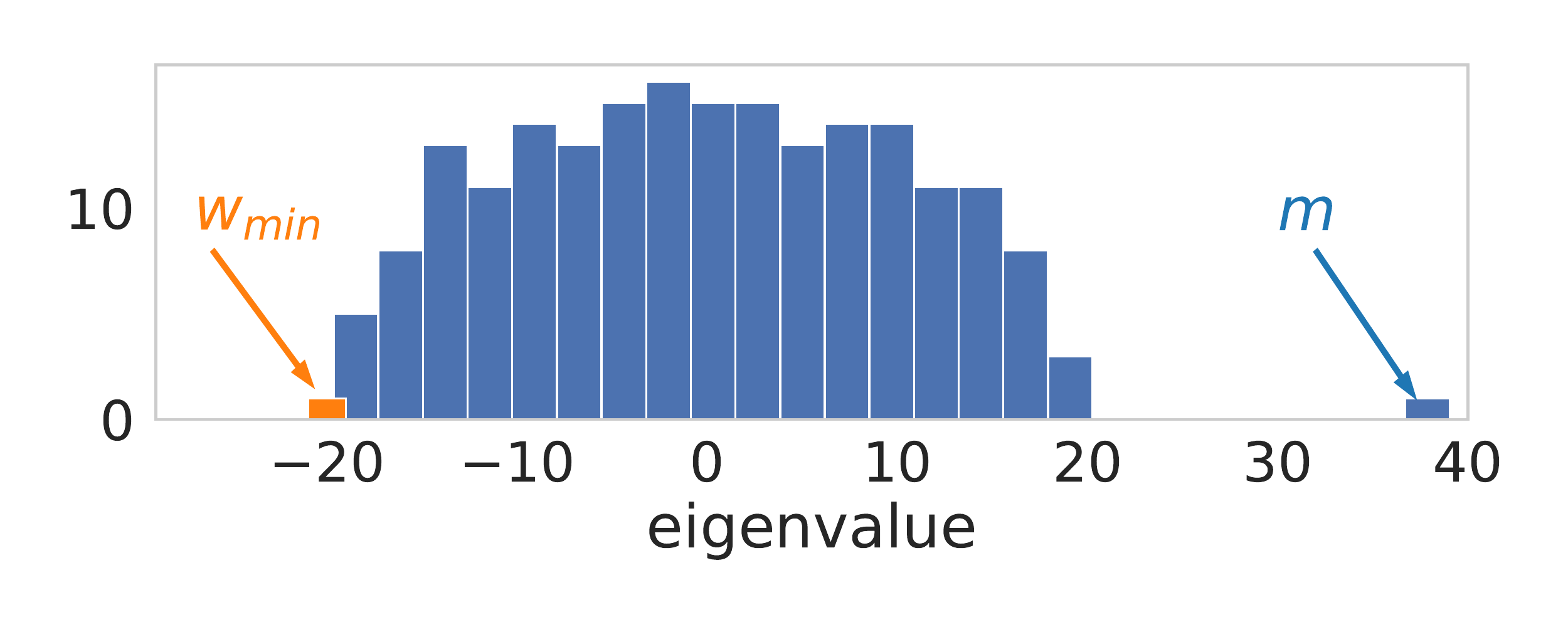}
    \label{fig:Per3}
    \caption{}
\end{subfigure}

\begin{subfigure}{0.9\textwidth}
    \centering
    \includegraphics[width=\textwidth]{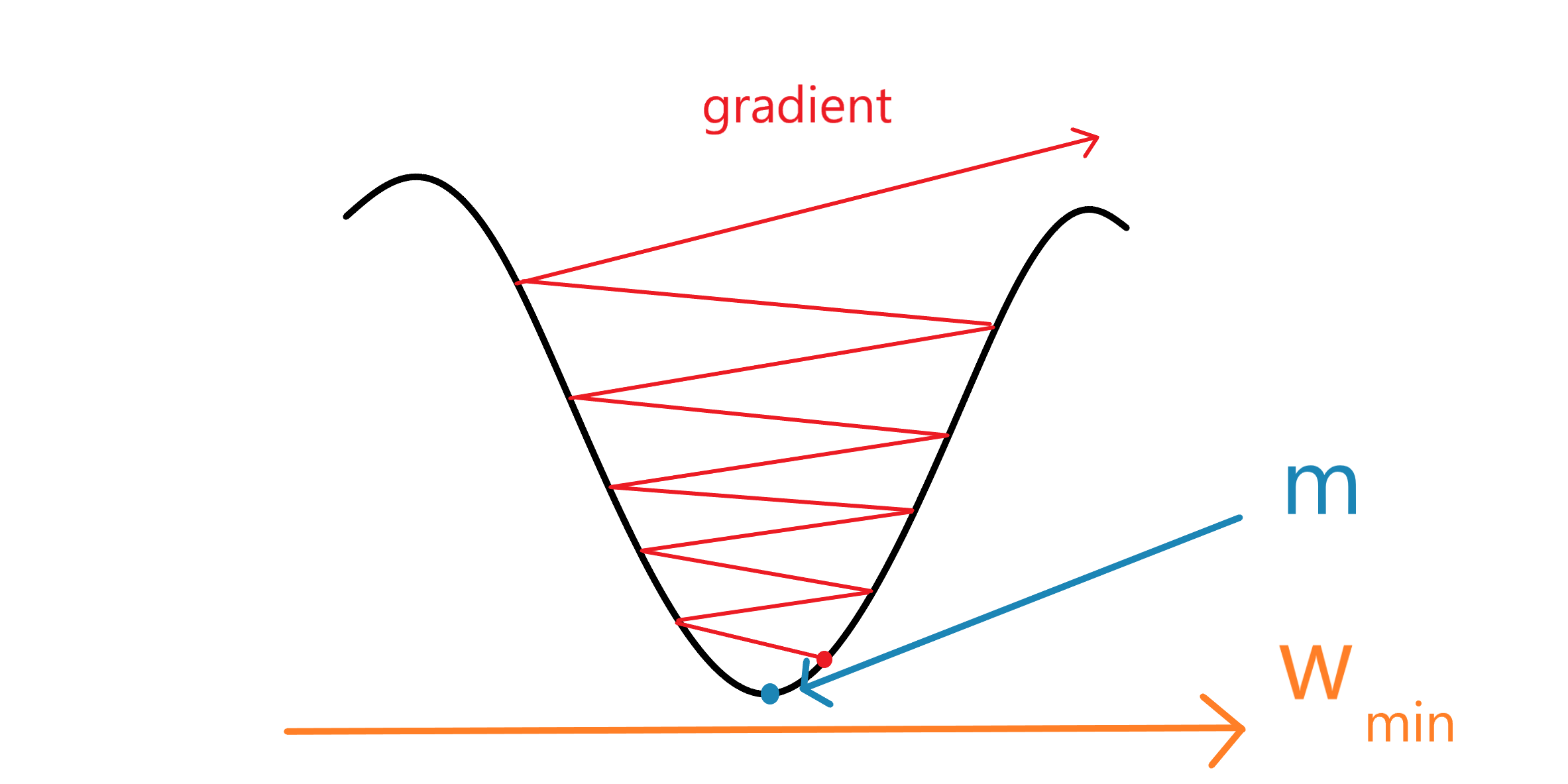}
    \label{fig:Per4}
    \caption{}
\end{subfigure}

\end{minipage} }
\caption{In (a) we observe that the algorithm gets stuck temporiraly in the basin of attraction of a local minimum $\vm_j$. $\vv_{i+1}-\vv_i$ becomes correlated to $w_\text{min}^j$ (the smallest eigenvector of $\tT (:,\vm_i,:)$: $w_\text{min}^j$ as illustrated in (b)) and its norm grows until the algorithm diverges away from $\vm_j$. This simple mechanism is illustrated in (c)  }
\label{escape}
\end{figure}

This simple first order approximation analysis is enough to capture the behavior of SMPI when escaping local minima. Indeed, we can see in the Figure \ref{escape}, a numerical example that illustrates this mechanism. $\vm_j$ denotes the closest local minimum whose basin of attraction is slowing the gradient descent. $\vw_\text{min}^j$ denotes the eigenvector associated to the smallest eigenvalue to the Hessian matrix at $\vm_j$. $\vv_i$ is the vector obtained after $i$ iterations on a random initialization. We see in Figure 4.a how SMPI gets stuck in a local minima $\vm_j$ for some time, we observe that at the same time $\displaystyle \langle \vv_{i+1}-\vv_i, \vw_\text{min}^j \rangle$ grows and $\displaystyle \langle \frac{\vv_{i+1}-\vv_i}{\norm{\vv_{i+1}-\vv_i}}, \vw_\text{min}^j \rangle$ becomes close to $1$. This illustrates an oscillating around the minimum $\vm_j$ along the axis corresponding to the minimal eigenvector $\vw_\text{min}^j$ of $\tT (:,\vm_i,:)$: $w_\text{min}^j$ and finally diverge away from it as pictured in Figure 4.c.

This exact same pattern happens in most of the initializations that succesfully converge towards the signal. In Table \ref{Tab:Escape}, we counted, for a successful initialization, the averaged number of spurious minima where the algorithm gets temporarily stuck before escaping thanks to its large step size. For different values of $n$, we repeated the experience for 20 independent instances and averaged the number of escaped minima before converging to the signal.

\begin{table}[h!]
\caption{We compute, for a successful initialization, the averaged number of spurious minima that we escape before converging towards the signal in function of $n$}
\label{sample-table5}
\begin{center} \begin{tabular}{c|c|c|c|c}
\hline
   $n$ &  $50$ &  $100$&  $150$& $200$ \\
   \hline
   number of minima that the algorithm escaped &  0.63 & 1.11 & 1.23 & 2.16 \\
   \hline
   
\end{tabular}
\end{center}
\label{Tab:Escape}
\end{table}

\subsubsection{The role of a polynomial iterations}


It is commonly assumed that Tensor power iteration functions by increasing the correlation with the signal at each step as in the matrix case.
\cite{richard2014statistical} performed a heuristic analysis with a zero order approximation that suggests that an initialization that verifies $\beta \langle \vv,\vv_0 \rangle>1$ is required for the method to succeed by increasing the correlation.

Yet, the results in Figure 4 left, where we plotted in red the initial correlation of successful vectors, strongly suggests that the success is not correlated with its initial correlation. Moreover Figure 4 right where we plot the correlation with the signal in function of the iteration shows that SMPI does not increase the initial correlation, but seems to rather have a first long phase of fluctuation of exploring the landscape. Indeed, the operating mode of the algorithm seems to be different when we consider moderately long times ($O(n)$). This indicates that this new mechanism is fundamentally different from what happens in the matrix case, and that a polynomial number of iterations is required.

Indeed a logarithmic convergence without exploration of the landscape would require a large initial correlation with the signal vector. Below the threshold $O(n^{1/2})$, this has an exponentially small probability to  happen  given  that  the distribution of the correlation of a random vector $\vv$ with the signal follows a normal law.

\begin{figure}
    \centering
\begin{subfigure}{0.5\textwidth}
    \centering
    \includegraphics[width=\textwidth]{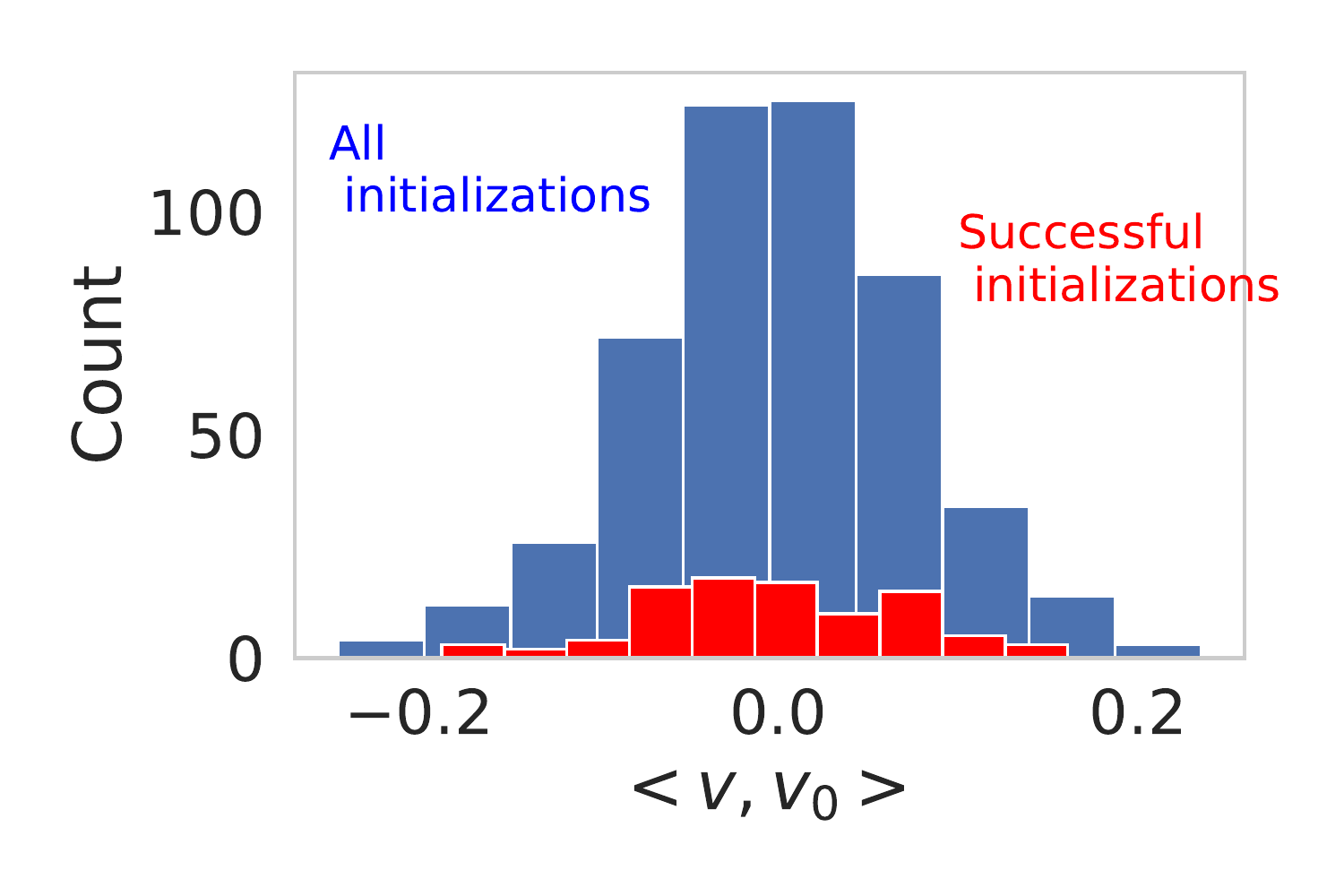}
    \label{fig:Per5}
\end{subfigure}%
\begin{subfigure}{0.5\textwidth}
    \centering
    \includegraphics[width=\textwidth]{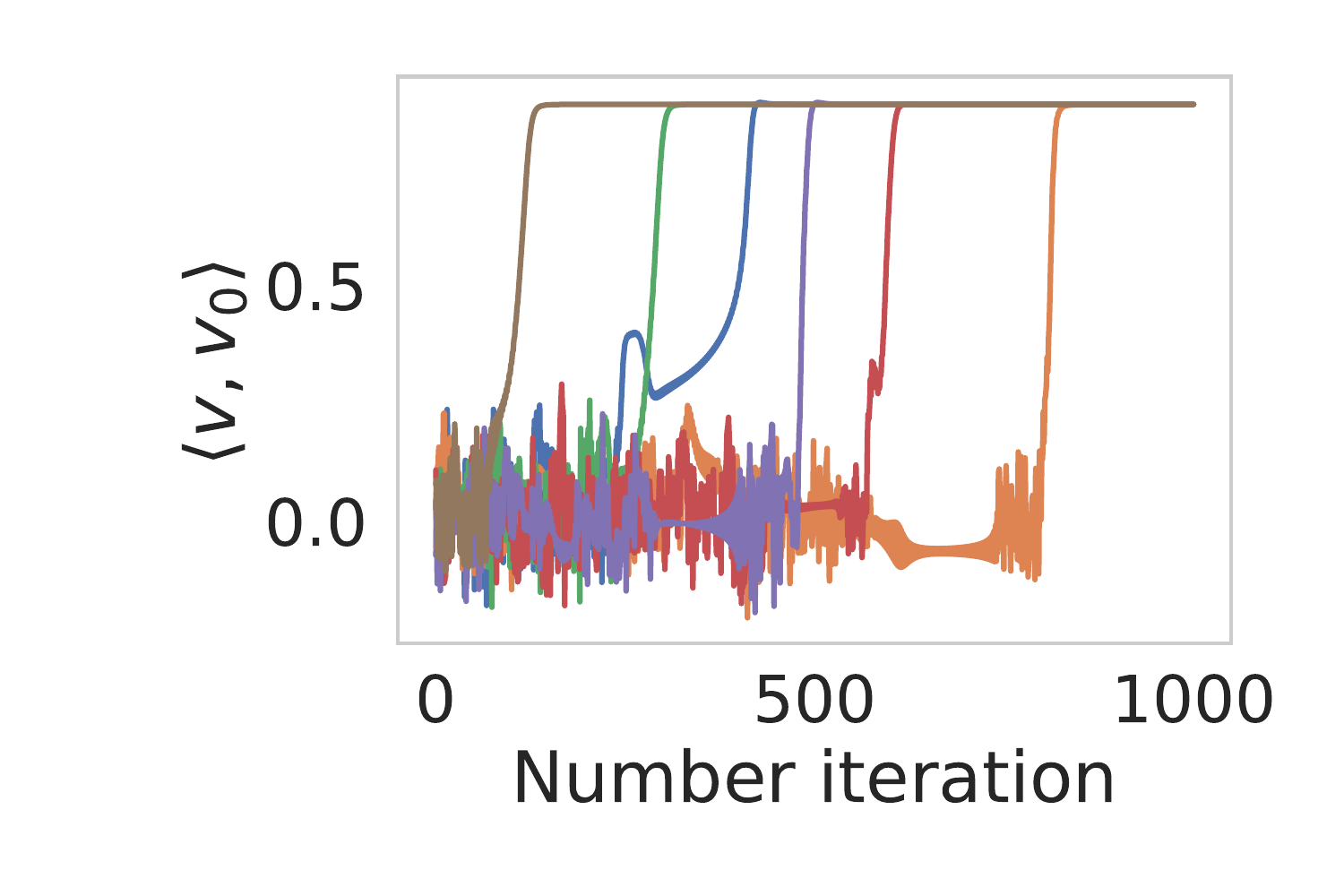}
    \label{fig:Per6}
\end{subfigure}
\caption{Left :  In blue the distribution of the correlation between the signal and all the initializations, and in red the correlations of the initializations that succeeded. We see that the initial correlation is not correlated to the final success. Right: Each color represents the trajectory of the algorithm for an initalization. We observe an initial phase of landscape exploration before converging towards the signal. These experiments are for a tensor with $n=200$, $m_{iter}=1000$ and $\beta=1.2$.}
\end{figure}


\subsubsection{The role of a polynomial initialisation}

In Table \ref{tab:polexp}, we reported in green the average of the number of initializations required to reach a success rate of $r=99\%$ with $m_{\text{iter}}=10*n$ over $10$ independent runs for each $n$ with SMPI. This is calculated by computing the percentage of successful initializations $p$ and then using the formula that gives the probability that at at least one of the initializations succeed: $1-(1-p)^{m_{99\%} }=0.99$. In red we reported the number of initializations required for the naive power initialization with logarithmic steps to succeed, which is exponential. 

\begin{table}[h!]
    \centering
\begin{tabular}{c|c|c|c|c}
   n & 50 & 100 & 200 & 400 \\
  \rowcolor{red!30} $\exp(n)\simeq$ & $ 10^{21}$ & $ 10^{43}$ & $ 10^{86}$ & $ 10^{173}$ \\
  \rowcolor{green!30}  $m_{init}$ - $99\% $  & 10 & 45 & 71 & 228  \\
\end{tabular}
    \caption{In green the average number of initializations required for a recovery rate success of $99\%$ where we see that it is linear in $n$. In red the approximation of $\exp(n)$ which is the number of required initializations if the complexity were exponential}
    \label{tab:polexp}
\end{table}

The drastic discrepancy between the two quantities suggest that SMPI has a polynomial complexity (and not an exponential complexity) even for $n>1000$.

\begin{figure}[h!]
    \centering
    \resizebox{0.7\textwidth}{!}{%
    \includegraphics[width=0.95\textwidth]{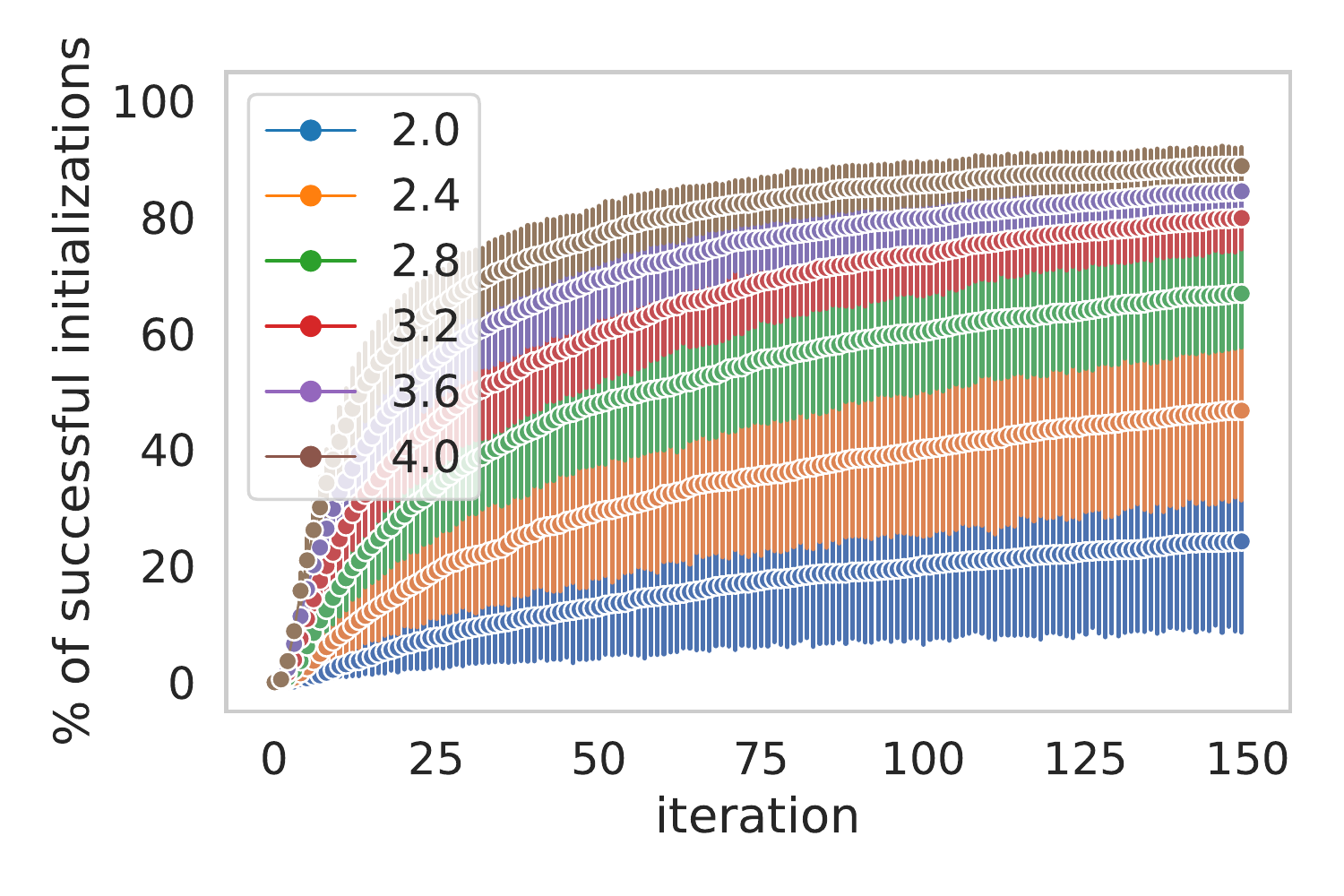}}
    \label{fig:max}

    \caption{Percentage of successful initializations in function of the number of iteration for different SNR $\beta$.}
    \label{fig:new}
\end{figure}


\subsection{Insight on the success}
    
Tensor power iteration with a random initialization is supposed to perform poorly with a computational threshold scaling as $n^{1/2}$ \cite{richard2014statistical} in contrast with other algorithms (such as unfolding, sum of squares, homotopy, etc.) whose algorithms thresholds scales as $n^{1/4}$. In order to understand the reason behind this failure, a first line of research focused on the complexity of the landscape (such as \cite{arous2019landscape} and \cite{ros2019complex}) showing the existence of an exponential number of spurious minima where the algorithm could get stuck. However, a more recent paper \cite{arous2020algorithmic} provided a proof for the failure of Langevin dynamics as well as gradient descent with infinitely small learning rate (the results use gradient flow) that suggest that the failure of local algorithms are "\textit{actually due to the weakness of the signal in the region of maximal entropy for the uninformative prior}". This means that the problem is not that the exponentially large number of minima, but rather that the signal is not strong enough when "started uniformly at random" in order to be able to attract the gradient.
    
It is thus very interesting to take advantage of the numerical analysis we performed on SMPI to understand how it would be able to bypass this these possible explanations to the failure of local iterative algorithms.
    (i) In the previous subsection, we observed that for the majority of the successful convergence towards the signal, the algorithm runs through many spurious minima but is still able to escape them thanks to its large step size. This provides a possible explanation in how SMPI is able to navigate through such a rough landscape in order to attain the signal vector.
    (ii) Numerical simulations for a low enough SNR and a large enough $n$ (e.g. $n \geq 100$) shows that for every successful convergence towards the signal, it is the gradient associated to the noise $\tZ (:,\vv,\vv)$ that not only triggers the convergence but also carries it. This mechanism that we exhibited is in total adequacy with the results of \cite{arous2020algorithmic} as it is consistent with the fact that the signal is indeed too small for its associated gradient to converge towards the signal by itself. However, our results bring a novel important element (which has never been considered before to the best of our knowledge) which is that the noise gradient is also able to play an crucial role in the convergence. Thus, thanks to this phenomenon, the smallness of the signal does not necessarily imply the failure of the algorithm.


\section{Potential impact and open questions}  

\subsection{Practical applications and Tensor decomposition}
SMPI has also the advantage to be simple, parallelizable and easy to generalize. In fact,we also proposed (in the supplementary material) two variants of  this  new  algorithm  to  deal  with  the  recovery of a spike with different dimensions and to deal with the multiple spikes recovery  (related  to  CP  tensor  decomposition problem). These proposed  algorithms  also  outperform  existent methods in this context which shows a huge potential impact of SMPI for practical applications.
To illustrate this point, we performed an experiment on an actual application in order to compare the performance of our algorithm with the state of the art in real life data.

The application we chose consists in the Hyperspectral images (HSI). In \cite{liu2012denoising}, the authors compared CP decomposition method based on the Alternating Least Square (ALS) algorithm with existent methods (two-dimensional filter and Tucker3 that is based on a Tucker decomposition) to denoise HSI. Their numerical results show that the CP decomposition model using the ALS algorithm performs better than other considered methods as a denoising procedure. 

In order to judge the performance of our algorithm, we perform the same experiment with the same estimator and compare it with the ALS algorithm using the Python TensorLy package \cite{kossaifi2016tensorly}. The hyperspectral image we use is the open source data given in \cite{miller2018above} that we normalize. It consists of a tensor of size $\bf{R}^{425\times 861 \times 475}$ where $425$ is the number of spectral bands and $865 \times 475$ is the spatial resolution.

\begin{table}[h!]
\caption{Comparison of methods}
\label{sample-table7}
\begin{center} \begin{tabular}{c|c|c|c|c}
\hline
   $n_2$ &  $150$ &  $200$&  $400$& $800$ \\
   \hline
   ALS (TensorLy)  &  43.58 & 54.24 & 60.58 & 66.53 \\
   SMPI & \bf{44.24} & \bf{54.53} & \bf{60.93} & \bf{66.91} \\
   \hline
   
\end{tabular}
\end{center}
\label{Tab:Hyper}
\end{table}

More details on Hyperspectral denoising and on the experiment could be found in the appendix.

   






\subsection{Insights on the gradient-based exploration of high-dimensional non-convex landscapes}

Many recent papers \cite{mannelli2019afraid,mannelli2019passed,mannelli2020marvels} utilized the landscape of Tensor PCA and its variant Matrix Tensor PCA in order to investigate the behavior of gradient descent in non convex high dimensional landscapes even in regimes where it should be hard. To the best of our knowledge, the mechanism we exhibit has not been considered before. This may be due to the fact that a large step size is required especially that our results suggest that this may be fundamentally important . Thus, this algorithm could bring a novel perspective in how the random landscape itself can play an important role in the convergence towards the signal. Furthermore, it pushes us to be careful when generalizing results obtained using gradient flow to results on gradient descent with large step size.



\subsection{Insights on the statistical-computational gap conjecture}    

\paragraph{Comparison with the predicted maximal theoretical results}

In \cite{jagannath2020statistical} gives an analytical expression for the asymptotic theoretical optimal correlation with the signal vector for $n \rightarrow \infty$ that is rigorously proven for $k$ even and $k\geq 6$ and conjectured to be true for all $k$.

\begin{figure}[h!]
    \centering
    \resizebox{0.7\textwidth}{!}{%
    \includegraphics[width=0.95\textwidth]{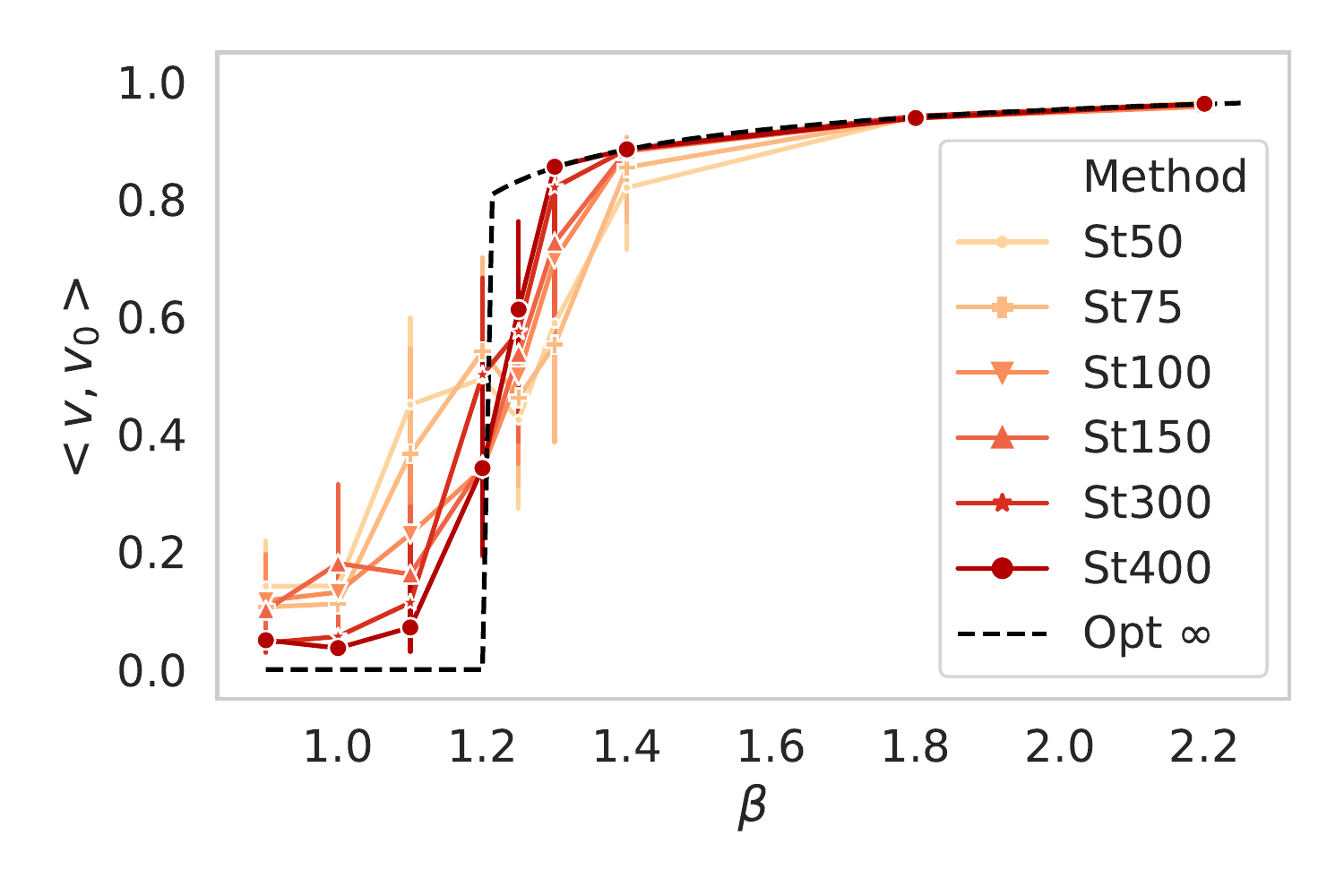}}
    \caption{Asymptotic behavior of SMPI method (named St in this figure) illustrated by different results on various values of $n$, the dimension of the axes of the tensor, ranging from 50 to 400. The dashed line (Opt $\infty$) corresponds to the predicted optimal theoretical result assuming $n=\infty$.}
    \label{fig:anc}
\end{figure}

Although the analytical formula is true only for $n \rightarrow \infty$, it is of interest to compare it with the performance of our algorithm for large $n$ like $n=50,75,100,150,300,400$. We plot in Figure \ref{fig:anc} the asymptotical curve in dashed line next to the results of SMPI for large $n$. We obtain a remarkable result: the empirical performance of SMPI converges towards the optimal performance for $n \rightarrow \infty$. Indeed the transition gets sharper when $n$ grows and its behavior gets closer to a discontinuous transition. Thus, these results suggest that not only the SMPI algorithm outperforms the state of the art but it may be matching the optimal information theory performance. In order to provide further information on what this algorithm performance implication on the statistical computational gap, one needs to investigate the complexity of the algorithm.

\subsubsection{Empirical scaling of the threshold}

To quantify the scaling of the thresholds of SMPI, we will make the assumption that the threshold can approximately be written as $\beta= c n^\alpha$ and our aim is to recover $\alpha$ empirically. Given $n_1$ and $n_2$, two values of $n$, we can show that the empirical $\alpha$ can be computed from the empirical thresholds as follows:
\begin{equation}
    \alpha_{\text{emp}} = \frac{\log (\beta_{n_2 }^{ \text{emp}})-\log (\beta_{n_1 }^{ \text{emp}})}{\log(n_2) - \log(n_1)} 
\end{equation}
In the following table we will take $n_1=100$ and vary $n_2$ in the set $\{150,200,400,800\}$. We perform 50 times the experiment for each $n_2$ and define $\beta^{\text{emp}}$ as the smaller $\beta$ where $\langle \vv,\vv_0 \rangle > 0.95  \langle \vv,\vv_0 \rangle_\text{th}$ where $\langle \vv,\vv_0 \rangle_\text{th}$ is the theoretical result for the maximum likehood estimator in the large $n$ limit given in \cite{jagannath2020statistical}.

\begin{table}[h!]
\caption{Experimental scaling for a non-symmetrical tensor for simple power iteration, unfolding method, homotopy-based method and SMPI}
\label{sample-table2}
\begin{center} \begin{tabular}{c|c|c|c|c}
\hline
   $n_2$ &  $150$ &  $200$&  $400$& $800$ \\
   \hline
   Simple power iteration &  0.541 & 0.528 & 0.531 & 0.513 \\
   Homotopy &  0.235 & 0.245 & 0.248 & 0.246 \\
   Unfolding &  0.23 & 0.248 & 0.26 & 0.2516 \\
   SMPI & -0.063 & -0.052 & -0.053 & -0.036 \\
   \hline
   
\end{tabular}
\end{center}
\label{Tab:comp2}
\end{table}

The first remark is that the two methods have an approximately constant $\alpha$, which confirms that the threshold behaves like $\beta= c n^\alpha$ in this range of $n$. This is in accordance with the references we gave in the previous answer where authors stated that the large $n$ behavior is observed starting from $n>50$.

\subsubsection{Discussion on a potential finite size effects}
\label{Finitesize}
It is logical and important to first consider the possibility that the experimental results that we obtained could be due to finite size effects. Therefore, we stress that the main aim and motivation of this work is not closing the gap but rather to provide novel theoretical and experimental insights that will help us understand better this conjectured gap either to prove it rigorously or to rule it out. 

First, it is important to note is that, for the best of our knowledge, all existent algorithms exhibits negligible finite size effects for $n>100$. Indeed, it is claimed that the empirical behavior of algorithms matches with the theoretical behavior for $n\geq 50$ for Unfolding and Tensor Power Iteration \cite{richard2014statistical}, $n \geq 30$ for Averaged Gradient Descent, $n \geq 100$ for Robust Tensor Power iteration \cite{wang2016online}, $n \geq 50$ for Higher Order Orthogonal Iteration of Tensors (HOOI) \cite{zhang2018tensor}. In our case, SMPI exhibits a constant threshold for $1 \leq n \leq 1000$.


Secondly, it could be interesting to discuss our results in light to a recent paper \cite{bandeira2020average} on the importance of considering finite size effects for low degree polynomial methods. Indeed, \cite{montanari2015non} observed that an SDP can tightly optimize a spiked matrix Hamiltonian. However, \cite{bandeira2020average} proved that it was actually due to finite size effect and the SDP is not tight. This emerges at around dimension $n\sim 10^4$. We discuss the fundamental difference between our results and these observations

\begin{itemize}
    \item The final asymptotic true result and the conjectured result by \cite{montanari2015non} differ only by a constant factor ($\sqrt{2}$) (and is due to a slow convergence speed of $O(n^{-1/2})$ to the asymptotic value). In contrast, in our case, constant factors are not relevant and the difference between what was previously conjectured for power iteration and SMPI is fundamentally different and equal to a factor of $n^{1/2}$. This challenges the idea of a slow convergence speed similar to the one in \cite{bandeira2020average}. 
    \item In the model of \cite{bandeira2020average}, the empirical performance stagnates before decreasing to the true asymptotic value as $n$ increases. This decrease of performance as $n$ increases is very typical of finite size effects. In contrast, we observe that as $n$ grows, the average correlation obtained by SMPI actually improves and converges towards the theoretical result obtained in the regime $n \rightarrow \infty$.
    \item The size of the tensor axis of $\tT$ and of a matrix axis of $\mM$ can't be directly compared. For example, there is more random variables in a tensor with $n_1=10^3$ than in a matrix with $n_2=10^4$ given that $n_1^3>n_2^2$. Thus, updating each component of the vector $\vv$ with a power iteration for $\vv \leftarrow \mM.\vv$ (where $\mM$'s axis is equal to $n_2$) consists in a sum of $10^4$ (pondered) random gaussian variables, while $\vv \leftarrow \tT(:,\vv, \vv)$  (where $\tT$'s axes dimension are equal to $n_1$) consists in a sum of $10^6$ (pondered) random variables. Therefore, intuition from Central limit theorem suggests that a tensor with $n_1=10^3$ should have less finite size effects than a matrix with $n_2=10^4$. Finally, one should keep in mind that $n=10^5$ for a tensor will require a storage capacity of 8 Petabytes that is 100 times larger than the best high-memory offers of cloud computing. However, this is not necessarily bothering since, comparing to the matrix case, the number of variables in a tensor is of order $n^3$, which leads to a rapid decrease of small size effects as it can be observed in the previously cited works.

\end{itemize}

\section{Conclusion}
In this paper, we introduced a novel algorithm named Selective Multiple Power Iteration (SMPI) for the important Tensor PCA problem. Various numerical simulations for $k=3$ in the conventionally considered range $n \leq 1000$ show that the experimental performance of SMPI improves drastically upon existent algorithms and becomes comparable to the theoretical optimal recovery. We also provide in the supplementary material multiple variants of this algorithm to tackle low-rank CP tensor decomposition. These proposed algorithms also outperforms existent methods even on real data which shows a huge potential impact for practical applications. 

Thus, for future work, it seems very interesting to go further in terms of theoretical investigations of these new insights offered by SMPI and also their consequences for related problems: the study of the behavior of gradient descent methods for the optimization in high-dimensional non-convex landscapes that are present in various machine learning problems and also the study of the conjectured statistical-algorithmic gap.

\section*{Acknowledgments}
The authors thank Gerard Ben Arous, Oleg Evnin, Reza Gheissari, Jiaoyang Huang and Aukosh Jagannath   for interesting and valuable discussions.


\bibliographystyle{unsrt}

\input{SMPI.bbl}
\newpage
\appendix


\section{ SMPI variants for generalizations of Tensor PCA}

\subsection{Algorithm for spike with different dimensions}
In order to make these tools versatile, less restrictive and usable for a majority of applications, we have to consider the case where the dimensions are different: $n_1 \neq n_2 \neq n_3. $. For example in a video, there is no reason to impose that the time dimension is equal to the two spatial dimensions.  Thus the problem is to infer a spike in the form of $\vv_1\otimes \vv_2 \otimes \vv_3$ from the following tensor
\begin{equation}
\displaystyle
    \tT=\beta \left(\frac{n_1+n_2+n_3}{3}\right)^{1/2} \vv_1\otimes \vv_2 \otimes \vv_3+\tZ,
    \label{PCADiffDim}
\end{equation}
where $\tZ \in \mathbb{R}^{n_1 \otimes n_2 \otimes n_3}$ is a tensor with random gaussian entries. Algorithm \ref{alg:DifferentDimensions} is an adaptation of SMPI to tackle this model.

\begin{algorithm}[h!]
   \caption{Recovery algorithm for a spike with different dimensions}
    \label{alg:DifferentDimensions}
\begin{algorithmic}[1]\onehalfspacing
   \STATE {\bfseries Input:} The tensor $\tT=\beta \left(\frac{n_1+n_2+n_3}{3}\right)^{1/2} \vv_a \otimes \vv_b \otimes \vv_c + \tZ $, $m_\text{init}$, $m_\text{iter}$
   \STATE {\bfseries Goal:} Estimate $\vv_a,\vv_b,\vv_c$.
   \STATE Generate $m_\text{init}$ random vectors $\{\vv_i\}_{ 1 \leq i \leq n_1 }$ and initialize ${\vv_a}_i^0={\vv_b}_i^0={\vv_c}_i^0=\vv_i$;
   \STATE Perform $m_\text{iter}$ times power iteration: 
   \begin{equation}
       \begin{split}
           {\vv_a}_{i}^{j+1} &\leftarrow \tT (:,{\vv_b}_i^{j}, {\vv_c}_i^{j}) / \norm{\tT (:,{\vv_b}_i^{j}, {\vv_c}_i^{j})}\\
           {\vv_b}_{i}^{j+1} &\leftarrow \tT ({\vv_a}_i^{j+1},:, {\vv_c}_i^{j}) / \norm{\tT ({\vv_a}_i^{j+1},:, {\vv_c}_i^{j})}\\
           {\vv_c}_{i}^{j+1} &\leftarrow \tT ({\vv_a}_i^{j+1}, {\vv_b}_i^{j+1},:) / \norm{\tT ({\vv_a}_i^{j+1}, {\vv_b}_i^{j+1},:)}
       \end{split}
   \end{equation}
   
   \STATE Select the vectors $({\vv_a}_f,{\vv_b}_f,{\vv_c}_f)=\arg \max_{1\leq i \leq {m_\text{init}}} \tT({\vv_a}^{m_\text{iter}}_i ,{\vv_b}^{m_\text{iter}}_i ,{\vv_c}^{m_\text{iter}}_i)$ :
   \STATE {\bfseries Output:} Obtaining estimated vectors $({\vv_a}_f,{\vv_b}_f,{\vv_c}_f)$
\end{algorithmic}
\end{algorithm}


\subsection{Low-rank CP decomposition algorithm}
We consider a generalization of the tensor PCA where we consider the problem of estimating multiple signal vectors. In this case, we can write the symmetric tensor with multiple spikes as:
\begin{equation}
\displaystyle
    \tT= \sum_{l=1}^p \sqrt{n} \beta_l \vv_l^{\otimes 3}+\tZ  \;\;\;\; 
\end{equation}

The algorithm \ref{alg:CPalg} is a simple variant of SMPI. 
Although it shares some similarities with existing methods based on power iteration such as \cite{wang2016online}, it is fundamentally different in its operating mode. Indeed, it is to the best of our knowledge, the first algorithm that is able to take advantage of the noise-based convergence mechanism given that it shares the same essential features with SMPI. This difference is illustrated by the substantial improvement over existing algorithms on our numerical experiments that we present in the next section.

\begin{algorithm}[h!]
   \caption{Recovery algorithm for for CP decomposition}
   \label{alg:CPalg}
\begin{algorithmic}[1]\onehalfspacing
   \STATE {\bfseries Input:} The tensor $\tT=\sum_{l=1}^p \sqrt{n} \beta_l \vv_l^{\otimes 3}+\tZ$, $m_\text{init}$, $m_\text{iter}$,$\varepsilon$,$\Lambda$($\sim n$)
   \STATE {\bfseries Goal:} Estimate $\{ \vv_l \}_{l=1}^p$ and $\{ \beta_l \}_{l=1}^p$.
   \STATE $\mathcal{E}=\emptyset$
   \STATE $\tT^0=\tT$
   
   \FOR{ i=0 to $m_\text{init}$} 
   \STATE Generate a random vector $\vv_{i,0}$
   \FOR{ j=0 to $m_\text{iter}$} 
   \STATE $\displaystyle\vv_{i,j+1}=\frac{\tT^i(:,\vv_{i,j},\vv_{i,j})}{\norm{\tT^i(:,\vv_{i,j},\vv_{i,j}) }}$
   \ENDFOR
   
   \IF{ $\displaystyle \norm{\vv_{i,m_\text{iter}-\Lambda},\vv_{i,m_\text{iter}}}\geq 1-\varepsilon $}
   \STATE 
      $\displaystyle\alpha_i=\langle \tT^{i}, \vv_{i,m_\text{iter}}^{\otimes k}\rangle $ 
      \STATE      $\displaystyle\tT^{i+1}=\tT^{i}-\alpha_i \vv_{i,m_\text{iter}}^{\otimes k} $
\STATE      $\displaystyle\mathcal{E}=\mathcal{E}+\{\vv_{i,m_\text{iter}}\}$
\ELSE
\STATE $\displaystyle\tT^{i+1}=\tT^{i}$
\ENDIF
\ENDFOR
\STATE {\bfseries Output:} Estimated vectors $\{ \hat{\vv}_l \}_{l=1}^p=\text{argmax}_{\mathcal{E}' \subset \mathcal{E}, |\mathcal{E}'|=p} \sum_{\vv \in \mathcal{E}'} \tT(\vv,\vv,\vv)$ and  $\hat{\beta}_l \equiv \alpha_l/\sqrt{n}$
   
\end{algorithmic}
\end{algorithm}


\section{Numerical simulations details}
Simulations for the comparison between SMPI and the unfolding method for $n=1000$ were run on a cloud provider (AWS). All the other simulations were run in Python on a Dell computer running Ubuntu 18.04.5 LTS with eight Intel Core i7-4800MQ processors at 2.70 GHz and 16GB of RAM.

\subsection{Spike with different dimensions}

\begin{figure}[h!]
    \centering
    \begin{subfigure}{.5\textwidth}
    \includegraphics[width=\columnwidth]{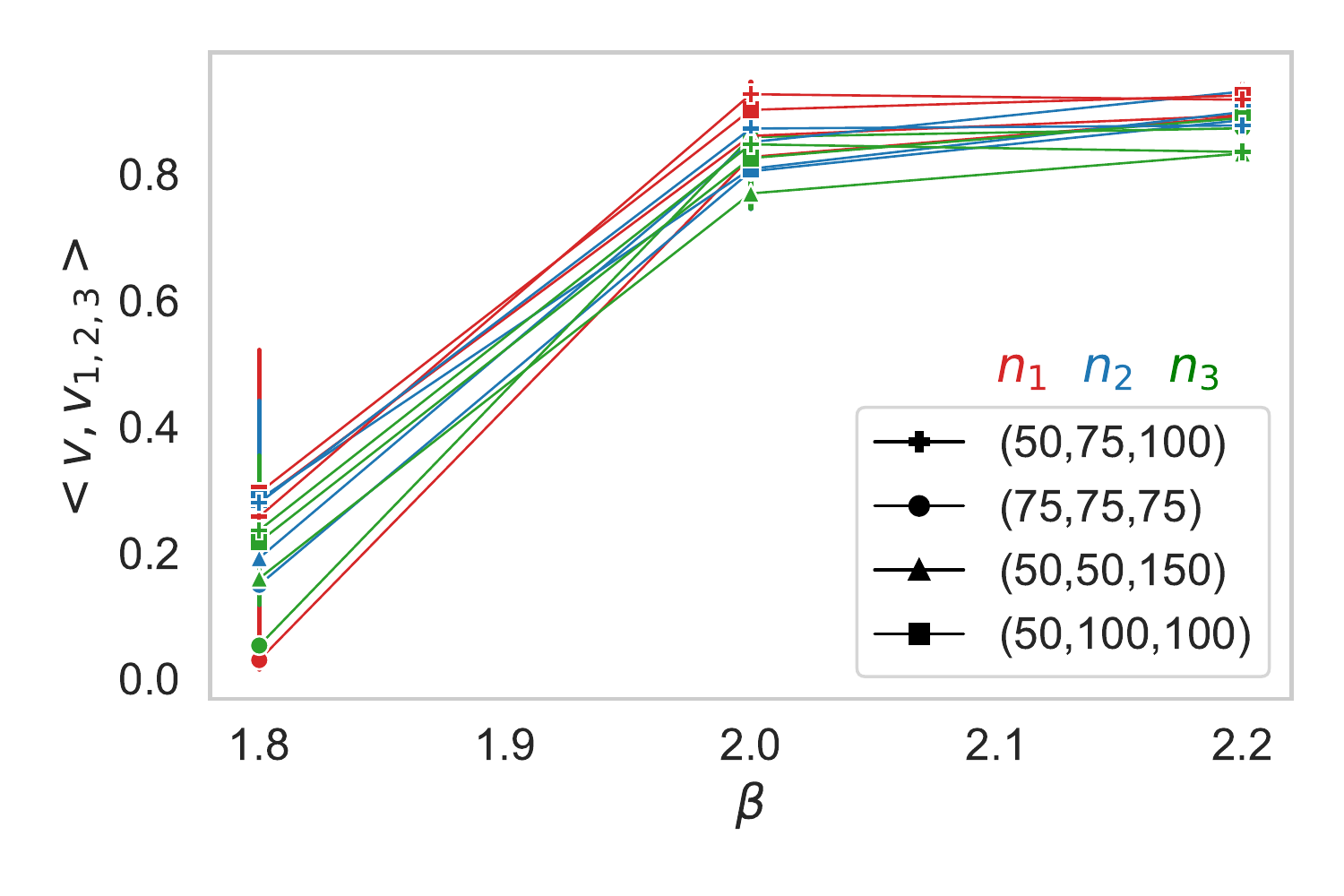}
\end{subfigure}
    \caption{ For different sets of three dimensions $(n_1,n_2,n_3)$, we generate different realizations of $\tT=\beta \left(\frac{n_1+n_2+n_3}{3}\right)^{1/2} \vv_1\otimes \vv_2 \otimes \vv_3+\tZ$ where $\tZ$ is a gaussian random tensor. We plot the correlation between the output of Algorithm \ref{alg:DifferentDimensions} and the signal vector (in red for $\langle \hat{\vv_1},\vv_1 \rangle$, in blue for $\langle \hat{\vv_2},\vv_2 \rangle$ and in green for $\langle \hat{\vv_3},\vv_3 \rangle$) in function of $\beta$.}
    \label{fig:DiffDim}
\end{figure}

In Figure \ref{fig:DiffDim}, we investigate the case of recovering a spike $\vv_1\otimes \vv_2 \otimes \vv_3$ with different dimensions $\tT=\beta \left(\frac{n_1+n_2+n_3}{3}\right)^{1/2} \vv_1\otimes \vv_2 \otimes \vv_3+\tZ$. For different sets of three dimensions $\{(50,75,100),(75,75,75),(50,50,150),(50,100,100)\}$ we plot the correlation between the outputs of the algorithm and the signal vectors (in red $\langle \hat{\vv_1},\vv_1 \rangle$, in blue $\langle \hat{\vv_2},\vv_2 \rangle$ and in green $\langle \hat{\vv_3},\vv_3 \rangle$) averaged over 50 different realizations. We see that the empirical threshold matches $(n_1+n_2+n_3)^{1/2}$ which corresponds to the optimal theoretical threshold for a tensor with dimension $n_1 \otimes n_2 \otimes n_3$.

\subsection{Multiple spikes case}

\begin{figure}[h!]
    \centering
    \begin{subfigure}{.45\textwidth}
    \includegraphics[width=\columnwidth]{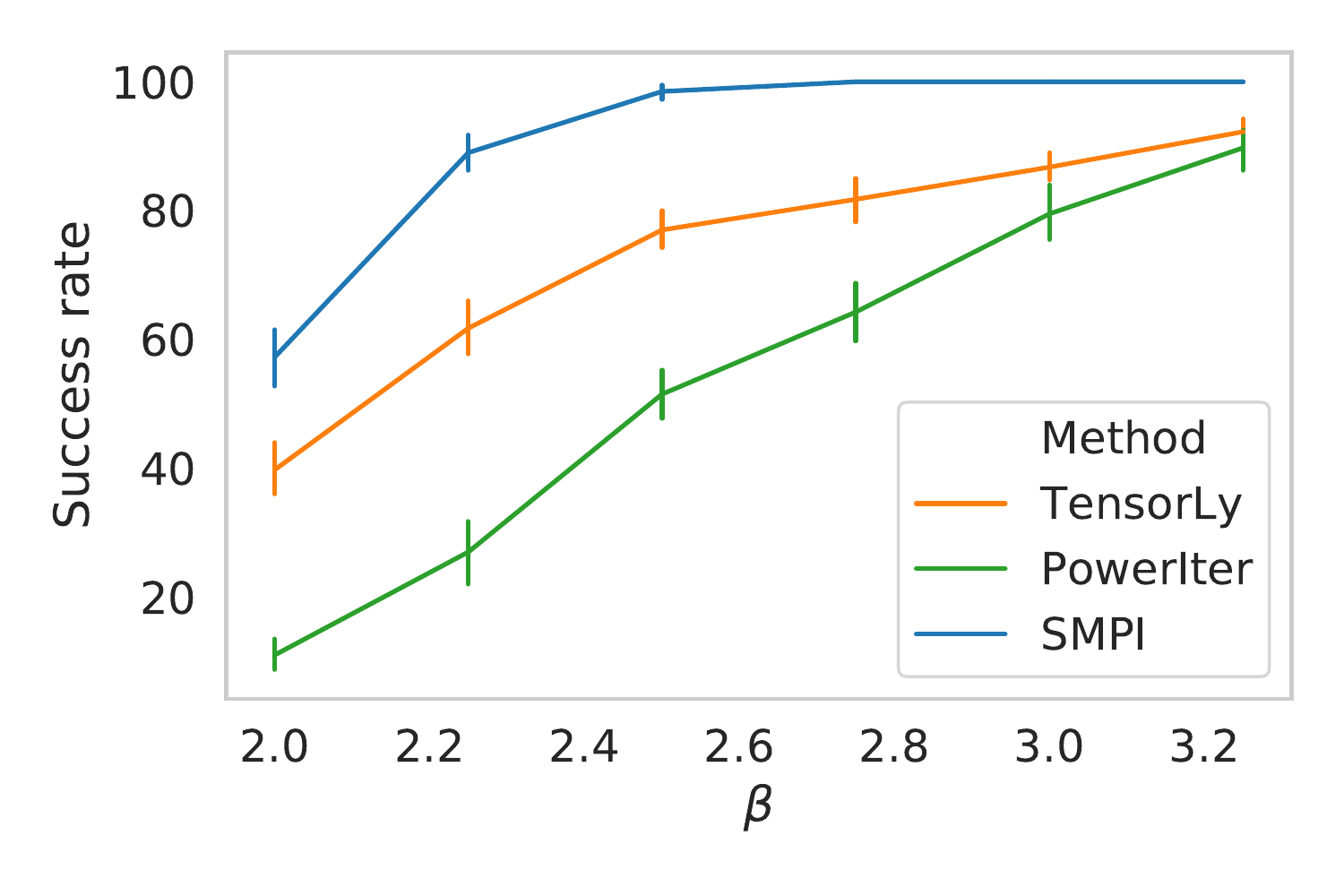}
    \label{fig:ManyItera}
    \end{subfigure}%
    \begin{subfigure}{.45\textwidth}
    \includegraphics[width=\columnwidth]{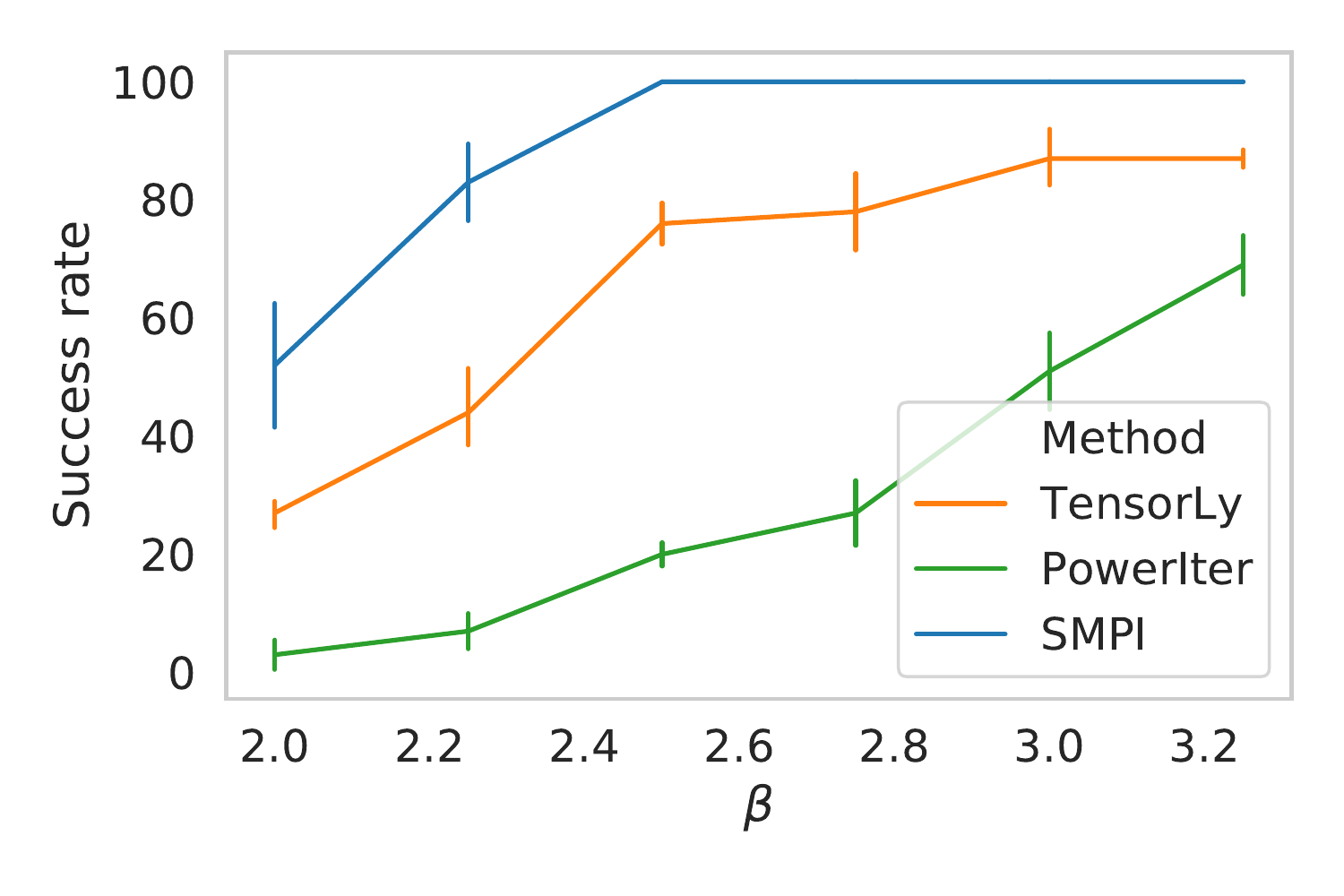}
    \label{fig:Four2}
    \end{subfigure}
    \caption{For a number of spikes equal to $20$, we plot the percentage of recovered spikes for different $\beta$ for $n=100$ in the left and $n=150$ in the right. We see that SMPI (blue) outperforms the naive power iteration algorithm (green) and the TensorLy algorithm (orange)}
    \label{cpdec}
\end{figure}

We investigate in Figure \ref{cpdec} the performance of the variant of SMPI for CP decomposition. We take $\beta$ equal for all the spikes since it is considered as the most difficult case for spectral algorithms. 

We repeat 30 times the following instance:
\begin{itemize}
    \item Generate the tensor $\tZ$ with iid Gaussian components
    \item Generate $n_\text{spikes}$ independent unitary random vectors. Each of these vectors is obtained by generating a vector with iid Gaussian components and then normalizing it.
    \item Compute $\tT=\tZ+\beta \sum_{l=1}^m \vv_{l} $
    \item Plot the percentage of successfully recovered vectors by algorithm \ref{alg:CPalg}, naive power iteration and the CP decomposition algorithm provided by the package TensorLy \cite{TensorLy:v20:18-277}.
\end{itemize}

In Figure \ref{cpdec}, we compare the percentage of succesfully recovered vectors with a naive power iteration and the CP decomposition algorithm provided by the package TensorLy \cite{TensorLy:v20:18-277}. We see that our algorithm outperforms existing methods.

\begin{figure}
    \centering
    \begin{subfigure}{.45\textwidth}
    \includegraphics[width=\columnwidth]{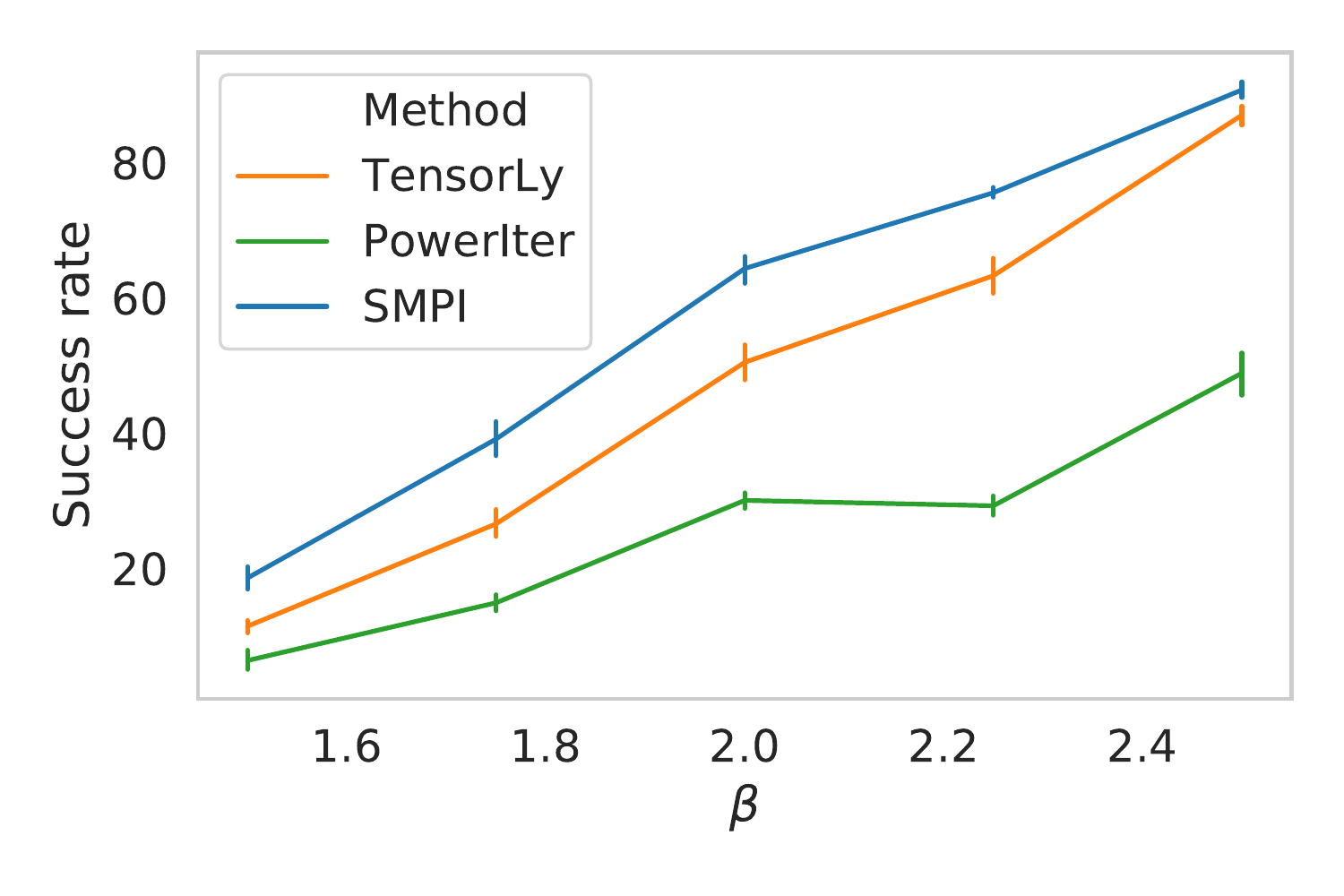}
    \label{fig:ManyItera2}
    \end{subfigure}
    \caption{For a number of spikes equal to 150, we plot the percentage of recovered spikes in function of $\beta$ averaged over 50 different tensors $\tT$ with $n =100$}
    \label{CPbetas}
\end{figure}

We investigate in Figure \ref{CPbetas} the case of a number of spikes larger than the dimension $n_\text{spikes} > n$. We see that even if the number of spikes is larger than the dimension, the algorithm still outperforms other methods.


\begin{figure}[h!]
    \centering
    \begin{subfigure}{.45\textwidth}
    \includegraphics[width=\columnwidth]{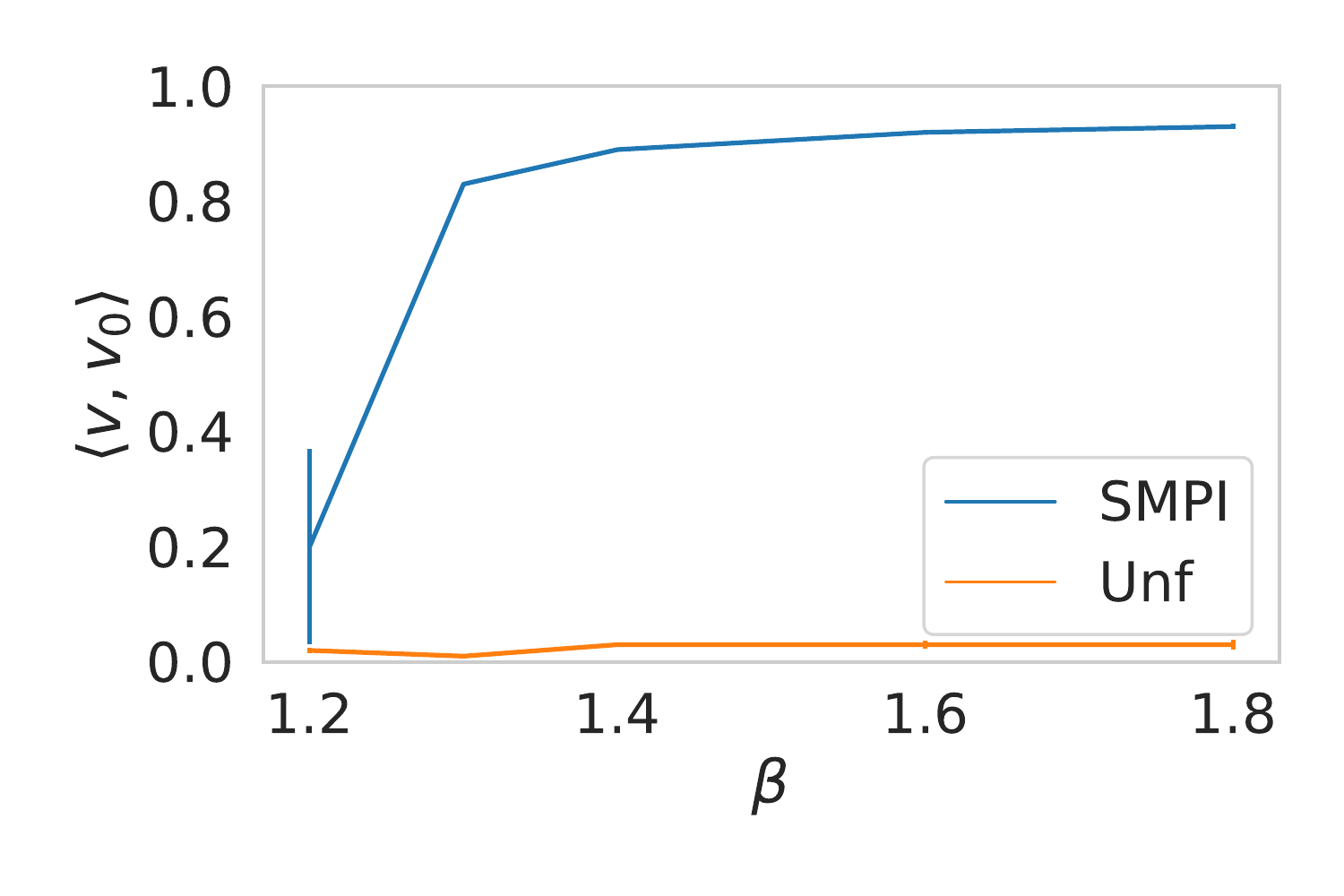}
    \caption{Comparison between the unfolding method and SMPI for $n=1000$ for different $\beta$}
    \label{fig:ManyItera3}
    \end{subfigure}\hfill
    \begin{subfigure}{.45\textwidth}
    \includegraphics[width=\columnwidth]{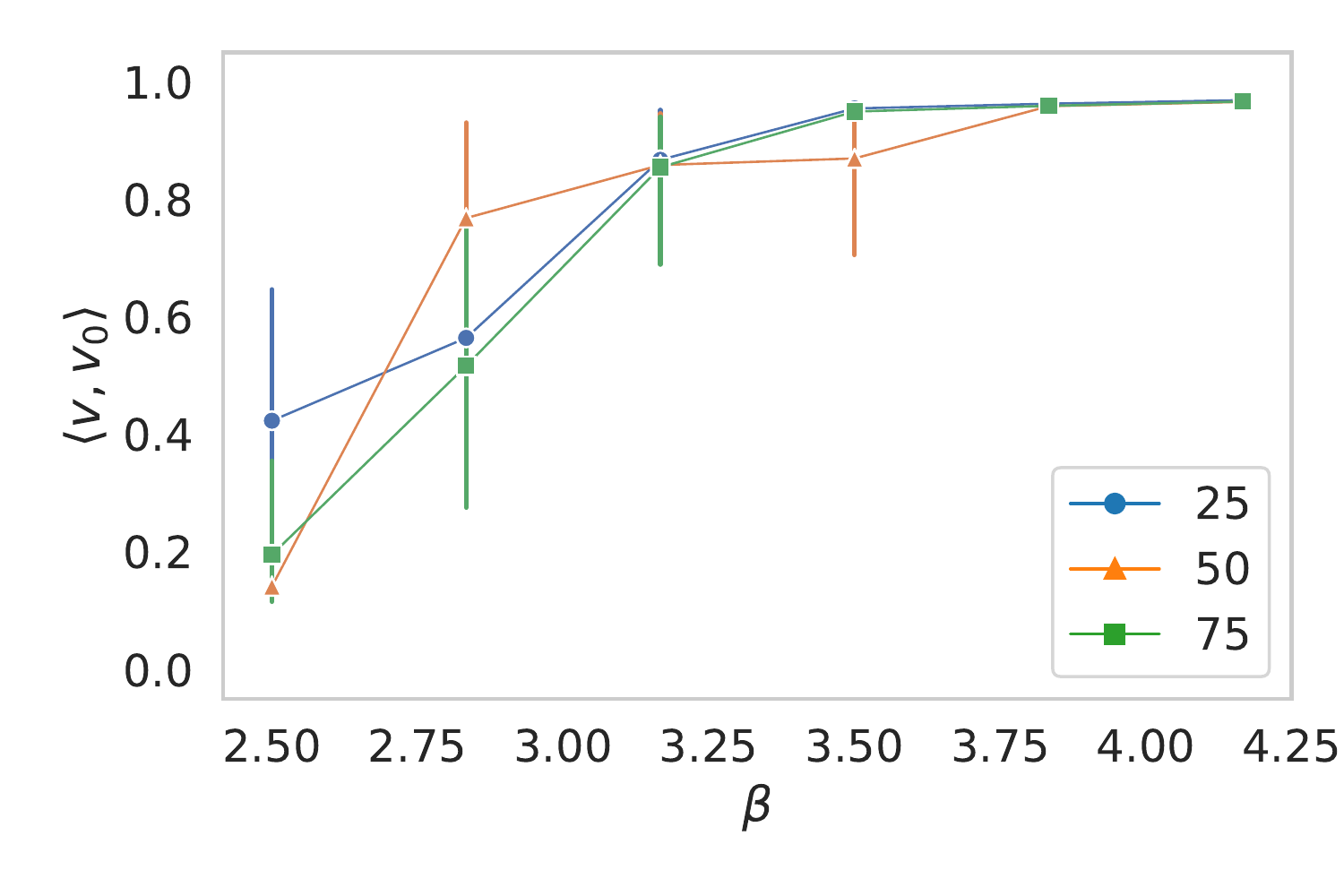}
    \caption{Correlation of the SMPI output $\vv$ with the signal $\vv_0$ for different $n \in \{25,50,75\}$ for a tensor of order $4$: $\tT \in \mathbb{R}^{n^{\otimes 4}}$}
    \label{fig:Four}
    \end{subfigure}
    \caption{The case $n=1000$ in the left and the performance of SMPI in the case of $k=4$ in the right}
\end{figure}

\section{Practical applications}

The application we chose consists in the Hyperspectral images (HSI). As explained in \cite{nasrabadi2013hyperspectral}, "Typically, a hyperspectral spectrometer provides hundreds of narrow contiguous bands over a wide range of the electromagnetic spectrum. Hyperspectral sensors measure the reflective (or emissive) properties of objects in the visible and short-wave infrared (IR) regions (or the mid-wave and long-wave IR regions) of the spectrum. Processing of these data allows algorithms to detect and identify targets of interest in a hyperspectral scene by exploiting the spectral signatures of the materials". 

Denoising is an important preprocessing step to further
analyze a hyperspectral image (HSI). The common denoising procedures are either based on 2-dimensional (2D) methods (mainly 2D filters) or tensor decomposition methods.
In particular, there is in general two mainly used models for tensor decomposition: Tucker decomposition and CP decomposition (also called PARAFAC) that have different advantages and shortcomings. A detailed survey of these methods could be found in \cite{kolda2009tensor} for example.

In \cite{liu2012denoising}, the authors compared CP decomposition method based on the Alternating Least Square (ALS) algorithm with existent methods (two-dimensional filter and Tucker3 that is based on a Tucker decomposition) to denoise HSI. 
They performed their experiments on a real world data: the Airborne Visible/Infrared Imaging Spectrometer (AVIRIS) HSI, an airborne hyperspectral system flown by NASA/Jet Propulsion
Laboratory (JPL). To analyze quantitatively the denoising results, they compare the signal-to-noise ratio (SNR) of the denoised image that is defined as:
\begin{equation}
    \text{SNR}_\text{out}=10 \log_{10} \frac{||\hat{\mathcal{Y}}||^2}{||\mathcal{Y}-\hat{\mathcal{Y}}||^2}
\end{equation}
where $\mathcal{Y}$ is the original image and $\hat{\mathcal{Y}}$ the estimated image after denoising. 

Their numerical results show that the CP decomposition model using the ALS algorithm performs better than other considered methods as a denoising procedure. 

In order to judge the performance of our algorithm, we perform the same experiment with the same estimator and compare it with the ALS algorithm using the Python TensorLy package \cite{kossaifi2016tensorly}. The hyperspectral image we use is the open source data given in \cite{miller2018above} that we normalize. It consists of a tensor of size $\bf{R}^{425\times 861 \times 475}$ where $425$ is the number of spectral bands and $865 \times 475$ is the spatial resolution.

CP decomposition model decomposes a tensor as a
sum of rank-one tensors (that we call spikes). Thus, we first compare the SNR for different number of spikes in Table \ref{sample-table3}.
Then we compare the time taken for each algorithm (in seconds) in Table \ref{sample-table4}.

\begin{table}[h!]
\caption{Comparison between ALS based on TensorLy and SMPI}
\label{sample-table3}
\begin{center} \begin{tabular}{c|c|c|c|c}
\hline
   $n_\text{spikes}$ &  $150$ &  $200$&  $400$& $800$ \\
   \hline
   ALS (TensorLy)  &  43.58 & 54.24 & 60.58 & 66.53 \\
   SMPI & \bf{44.24} & \bf{54.53} & \bf{60.93} & \bf{66.91} \\
   \hline
   
\end{tabular}
\end{center}
\label{Tab:Hyper4}
\end{table}

\begin{table}[h!]
\caption{Time for each method}
\label{sample-table4}
\begin{center} \begin{tabular}{c|c|c|c|c}
\hline
  $n_\text{spikes}$ &  $150$ &  $200$&  $400$& $800$ \\
  \hline
  ALS (TensorLy)  &  {\bf{4}} & \bf{377} & 1566 & 2325  \\
  SMPI & 33 & 387 & {\bf{995}} & {\bf{1823}} \\
  \hline
   
\end{tabular}
\end{center}
\label{Tab:Hyper2}
\end{table}

We see that the proposed SMPI method gives better denoising results independently of the number of spikes. And we see that the ALS algorithm (using the TensorLy package) is faster for small number of spikes but becomes slower than SMPI for a larger number of spikes. Given that the optimal number of spikes is in general $>100$ (in \cite{liu2012denoising} the optimal was for 169 spikes), this suggests that SMPI gives better results in a shorter amount in time than ALS.

More generally, it is important to note that there is many more practical applications where CP decomposition and where SMPI could be an excellent candidate for improving existent performance. Here is a non-exhaustive list of such applications:

In telecommunication, CP decomposition is used for Tensor-based modulation \cite{decurninge2020tensor}. It used for "Massive random access, whereby a large number of transmitters communicate with a single receiver, constitutes a key design challenge for future generations of wireless systems."

An important other application is the convolutional neural networks compression: for example a CP decomposition method based on power iteration has been suggested in \cite{astrid2017cp}  that showed a significant reduction in memory and computation cost. Given that our method could be seen as a refinement of the simple power iteration method, we believe that it could be very interesting to try SMPI on these models.

In \cite{sun2020opportunities}, where different types of higher order data in manufacturing processes are described, and their potential usage is addressed using methods like CP tensor decomposition.

\section{Theoretical expression for the plateau in equation \ref{plateau}}
\label{Appplateau}
Let's consider $\vv$ the final minima obtained by the maximum likehood estimator, given that it is a minima of a symmetric tensor we have the equality
$$\frac{\tT(:,\vv,\vv)}{\norm{\tT(:,\vv,\vv)}}=\vv$$
Computing the scalar product with $\bf{v_0}$ of each side and using that $\tT=\tZ+\beta \vv_0^{\otimes 3}$ we have
$$\langle \tZ(:,\vv,\vv),\vv_0 \rangle + \beta \langle \vv,\vv_0 \rangle^2=\langle \vv,\vv_0 \rangle || \tT(:,\vv,\vv) ||$$
So:
$$\langle \tZ(:,\vv,\vv),\vv_0 \rangle =\langle \vv,\vv_0\rangle  (|| \tT(:,\vv,\vv) || -  \beta \langle \vv,\vv_0 \rangle)$$
On the other side, we also have 
$$ \norm{\tT(:,\vv,\vv)}=\norm{\tZ(:,\vv,\vv)+ \beta \langle \vv,\vv_0 \rangle^2 \vv_0} $$
$$ \norm{\tT(:,\vv,\vv)}^2=\norm{\tZ(:,\vv,\vv)}^2+ \beta^2 \langle \vv,\vv_0 \rangle^4 +2 \beta \langle \vv,\vv_0 \rangle^2 \langle \tZ(:,\vv,\vv), \vv_0 \rangle $$
Thus

$$\frac{\langle \tZ(:,\vv,\vv),\vv_0 \rangle}{|| \bf{\tZ(:,\vv,\vv)} ||} =  \frac{\langle \vv,\vv_0\rangle(|| \tT(:,\vv,\vv) || -  \beta \langle \vv,\vv_0 \rangle)}{\sqrt{\norm{\tT(:,\vv,\vv)}^2 - \beta^2 \langle \vv,\vv_0 \rangle^4-2 \beta \langle \vv,\vv_0 \rangle^2 \langle \tZ(:,\vv,\vv), \vv_0 \rangle}}$$

replacing the obtaining expression of $\langle \tZ(:,\vv,\vv), \vv_0 \rangle$ in the right side of the equation

$$\frac{\langle \tZ(:,\vv,\vv),\vv_0 \rangle}{|| \tZ(:,\vv,\vv) ||} =  \frac{\langle \vv,\vv_0\rangle(|| \tT(:,\vv,\vv) || -  \beta \langle \vv,\vv_0 \rangle)}{\sqrt{\norm{\tT(:,\vv,\vv)}^2 + \beta^2 \langle \vv,\vv_0 \rangle^4-2 \beta \langle \vv,\vv_0 \rangle^3 \norm{\tT(:,\vv,\vv)}}}$$

\end{document}

%% file: math_commands.tex
\usepackage{amsmath,amsfonts,bm}

\def\eqref#1{equation~\ref{#1}}

\def\1{\bm{1}}

\def\vg{{\bm{g}}}

\def\vm{{\bm{m}}}

\def\vv{{\bm{v}}}
\def\vw{{\bm{w}}}

\def\vy{{\bm{y}}}
\def\vz{{\bm{z}}}

\def\evv{{v}}
\def\evw{{w}}

\def\evz{{z}}

\def\mM{{\bm{M}}}

\DeclareMathAlphabet{\mathsfit}{\encodingdefault}{\sfdefault}{m}{sl}
\SetMathAlphabet{\mathsfit}{bold}{\encodingdefault}{\sfdefault}{bx}{n}
\newcommand{\tens}[1]{\bm{\mathsfit{#1}}}

\def\tT{{\tens{T}}}

\def\tZ{{\tens{Z}}}

\def\emM{{M}}

\newcommand{\etens}[1]{\mathsfit{#1}}

\def\etT{{\etens{T}}}








%% file: SMPI.bbl
\begin{thebibliography}{10}

\bibitem{richard2014statistical}
Emile Richard and Andrea Montanari.
\newblock A statistical model for tensor pca.
\newblock In {\em Advances in Neural Information Processing Systems}, pages
  2897--2905, 2014.

\bibitem{wang2016online}
Yining Wang and Animashree Anandkumar.
\newblock Online and differentially-private tensor decomposition.
\newblock In {\em Proceedings of the 30th International Conference on Neural
  Information Processing Systems}, NIPS'16, page 3539–3547, Red Hook, NY,
  USA, 2016. Curran Associates Inc.

\bibitem{anandkumar2015learning}
Animashree Anandkumar, Rong Ge, and Majid Janzamin.
\newblock Learning overcomplete latent variable models through tensor methods.
\newblock In {\em Conference on Learning Theory}, pages 36--112. PMLR, 2015.

\bibitem{anandkumar2013tensor}
Animashree Anandkumar, Rong Ge, Daniel Hsu, and Sham Kakade.
\newblock A tensor spectral approach to learning mixed membership community
  models.
\newblock In {\em Conference on Learning Theory}, pages 867--881. PMLR, 2013.

\bibitem{astrid2017cp}
Marcella Astrid and Seung-Ik Lee.
\newblock Cp-decomposition with tensor power method for convolutional neural
  networks compression.
\newblock In {\em 2017 IEEE International Conference on Big Data and Smart
  Computing (BigComp)}, pages 115--118. IEEE, 2017.

\bibitem{wang2020cpac}
Yinan Wang, Xiaowei Yue, et~al.
\newblock Cpac-conv: Cp-decomposition to approximately compress convolutional
  layers in deep learning.
\newblock {\em arXiv preprint arXiv:2005.13746}, 2020.

\bibitem{arous2020algorithmic}
Gerard~Ben Arous, Reza Gheissari, Aukosh Jagannath, et~al.
\newblock Algorithmic thresholds for tensor pca.
\newblock {\em Annals of Probability}, 48(4):2052--2087, 2020.

\bibitem{mannelli2019passed}
Stefano~Sarao Mannelli, Florent Krzakala, Pierfrancesco Urbani, and Lenka
  Zdeborova.
\newblock Passed \& spurious: Descent algorithms and local minima in spiked
  matrix-tensor models.
\newblock In {\em international conference on machine learning}, pages
  4333--4342. PMLR, 2019.

\bibitem{mannelli2019afraid}
Stefano~Sarao Mannelli, Giulio Biroli, Chiara Cammarota, Florent Krzakala, and
  Lenka Zdeborov{\'a}.
\newblock Who is afraid of big bad minima? analysis of gradient-flow in a
  spiked matrix-tensor model.
\newblock {\em arXiv preprint arXiv:1907.08226}, 2019.

\bibitem{mannelli2020marvels}
Stefano~Sarao Mannelli, Giulio Biroli, Chiara Cammarota, Florent Krzakala,
  Pierfrancesco Urbani, and Lenka Zdeborov{\'a}.
\newblock Marvels and pitfalls of the langevin algorithm in noisy
  high-dimensional inference.
\newblock {\em Physical Review X}, 10(1):011057, 2020.

\bibitem{luo2020open}
Yuetian Luo and Anru~R Zhang.
\newblock Open problem: Average-case hardness of hypergraphic planted clique
  detection.
\newblock In {\em Conference on Learning Theory}, pages 3852--3856. PMLR, 2020.

\bibitem{hopkins2016fast}
Samuel~B Hopkins, Tselil Schramm, Jonathan Shi, and David Steurer.
\newblock Fast spectral algorithms from sum-of-squares proofs: tensor
  decomposition and planted sparse vectors.
\newblock In {\em Proceedings of the forty-eighth annual ACM symposium on
  Theory of Computing}, pages 178--191, 2016.

\bibitem{homotopy17a}
Anima Anandkumar, Yuan Deng, Rong Ge, and Hossein Mobahi.
\newblock Homotopy analysis for tensor pca.
\newblock In Satyen Kale and Ohad Shamir, editors, {\em Proceedings of the 2017
  Conference on Learning Theory}, volume~65 of {\em Proceedings of Machine
  Learning Research}, pages 79--104. PMLR, 07--10 Jul 2017.

\bibitem{wein2019kikuchi}
Alexander~S Wein, Ahmed El~Alaoui, and Cristopher Moore.
\newblock The kikuchi hierarchy and tensor pca.
\newblock In {\em 2019 IEEE 60th Annual Symposium on Foundations of Computer
  Science (FOCS)}, pages 1446--1468. IEEE, 2019.

\bibitem{biroli2020iron}
Giulio Biroli, Chiara Cammarota, and Federico Ricci-Tersenghi.
\newblock How to iron out rough landscapes and get optimal performances:
  averaged gradient descent and its application to tensor pca.
\newblock {\em Journal of Physics A: Mathematical and Theoretical},
  53(17):174003, 2020.

\bibitem{Hastings2020classicalquantum}
Matthew~B. Hastings.
\newblock Classical and {Q}uantum {A}lgorithms for {T}ensor {P}rincipal
  {C}omponent {A}nalysis.
\newblock {\em {Quantum}}, 4:237, February 2020.

\bibitem{kunisky2019notes}
Dmitriy Kunisky, Alexander~S Wein, and Afonso~S Bandeira.
\newblock Notes on computational hardness of hypothesis testing: Predictions
  using the low-degree likelihood ratio.
\newblock {\em arXiv preprint arXiv:1907.11636}, 2019.

\bibitem{dudeja2021statistical}
Rishabh Dudeja and Daniel Hsu.
\newblock Statistical query lower bounds for tensor pca.
\newblock {\em Journal of Machine Learning Research}, 22(83):1--51, 2021.

\bibitem{Ouerfelli2022random}
Mohamed Ouerfelli, Mohamed Tamaazousti, and Vincent Rivasseau.
\newblock Random tensor theory for tensor decomposition.
\newblock In {\em Proceedings of the AAAI Conference on Artificial
  Intelligence}, 2022.

\bibitem{lahoche2021field}
Vincent Lahoche, Mohamed Ouerfelli, Dine~Ousmane Samary, and Mohamed
  Tamaazousti.
\newblock Field theoretical approach for signal detection in nearly continuous
  positive spectra ii: Tensorial data.
\newblock {\em Entropy}, 23(7):795, 2021.

\bibitem{anandkumar2012tensor}
Anima Anandkumar, Rong Ge, Daniel Hsu, Sham~M Kakade, and Matus Telgarsky.
\newblock Tensor decompositions for learning latent variable models.
\newblock {\em arXiv preprint arXiv:1210.7559}, 2012.

\bibitem{huang2020power}
Jiaoyang Huang, Daniel~Z Huang, Qing Yang, and Guang Cheng.
\newblock Power iteration for tensor pca.
\newblock {\em arXiv preprint arXiv:2012.13669}, 2020.

\bibitem{choromanska2015loss}
Anna Choromanska, Mikael Henaff, Michael Mathieu, G{\'e}rard~Ben Arous, and
  Yann LeCun.
\newblock The loss surfaces of multilayer networks.
\newblock In {\em Artificial intelligence and statistics}, pages 192--204.
  PMLR, 2015.

\bibitem{brennan2020reducibility}
Matthew Brennan and Guy Bresler.
\newblock Reducibility and statistical-computational gaps from secret leakage.
\newblock In {\em Conference on Learning Theory}, pages 648--847. PMLR, 2020.

\bibitem{perry2020statistical}
Amelia Perry, Alexander~S Wein, Afonso~S Bandeira, et~al.
\newblock Statistical limits of spiked tensor models.
\newblock In {\em Annales de l'Institut Henri Poincar{\'e}, Probabilit{\'e}s et
  Statistiques}, volume~56, pages 230--264. Institut Henri Poincar{\'e}, 2020.

\bibitem{lesieur2017statistical}
Thibault Lesieur, L{\'e}o Miolane, Marc Lelarge, Florent Krzakala, and Lenka
  Zdeborov{\'a}.
\newblock Statistical and computational phase transitions in spiked tensor
  estimation.
\newblock In {\em 2017 IEEE International Symposium on Information Theory
  (ISIT)}, pages 511--515. IEEE, 2017.

\bibitem{ros2019complex}
Valentina Ros, Gerard~Ben Arous, Giulio Biroli, and Chiara Cammarota.
\newblock Complex energy landscapes in spiked-tensor and simple glassy models:
  Ruggedness, arrangements of local minima, and phase transitions.
\newblock {\em Physical Review X}, 9(1):011003, 2019.

\bibitem{jagannath2020statistical}
Aukosh Jagannath, Patrick Lopatto, Leo Miolane, et~al.
\newblock Statistical thresholds for tensor pca.
\newblock {\em Annals of Applied Probability}, 30(4):1910--1933, 2020.

\bibitem{Hom2016arXiv161009322A}
Anima Anandkumar, Yuan Deng, Rong Ge, and Hossein Mobahi.
\newblock Homotopy analysis for tensor pca.
\newblock In {\em Conference on Learning Theory}, pages 79--104. PMLR, 2017.

\bibitem{TensorLy:v20:18-277}
Jean Kossaifi, Yannis Panagakis, Anima Anandkumar, and Maja Pantic.
\newblock Tensorly: Tensor learning in python.
\newblock {\em Journal of Machine Learning Research}, 20(26):1--6, 2019.

\bibitem{arous2019landscape}
Gerard~Ben Arous, Song Mei, Andrea Montanari, and Mihai Nica.
\newblock The landscape of the spiked tensor model.
\newblock {\em Communications on Pure and Applied Mathematics},
  72(11):2282--2330, 2019.

\bibitem{liu2012denoising}
Xuefeng Liu, Salah Bourennane, and Caroline Fossati.
\newblock Denoising of hyperspectral images using the parafac model and
  statistical performance analysis.
\newblock {\em IEEE Transactions on Geoscience and Remote Sensing},
  50(10):3717--3724, 2012.

\bibitem{kossaifi2016tensorly}
Jean Kossaifi, Yannis Panagakis, Anima Anandkumar, and Maja Pantic.
\newblock Tensorly: Tensor learning in python.
\newblock {\em arXiv preprint arXiv:1610.09555}, 2016.

\bibitem{miller2018above}
CE~Miller, RO~Green, DR~Thompson, AK~Thorpe, M~Eastwood, IB~Mccubbin,
  W~Olson-Duvall, M~Bernas, CM~Sarture, S~Nolte, et~al.
\newblock Above: Hyperspectral imagery from aviris-ng, alaskan and canadian
  arctic, 2017--2018.
\newblock {\em ORNL DAAC}, 2018.

\bibitem{zhang2018tensor}
Anru Zhang and Dong Xia.
\newblock Tensor svd: Statistical and computational limits.
\newblock {\em IEEE Transactions on Information Theory}, 64(11):7311--7338,
  2018.

\bibitem{bandeira2020average}
Afonso~S Bandeira, Dmitriy Kunisky, and Alexander~S Wein.
\newblock Average-case integrality gap for non-negative principal component
  analysis.
\newblock {\em arXiv preprint arXiv:2012.02243}, 2020.

\bibitem{montanari2015non}
Andrea Montanari and Emile Richard.
\newblock Non-negative principal component analysis: Message passing algorithms
  and sharp asymptotics.
\newblock {\em IEEE Transactions on Information Theory}, 62(3):1458--1484,
  2015.

\bibitem{nasrabadi2013hyperspectral}
Nasser~M Nasrabadi.
\newblock Hyperspectral target detection: An overview of current and future
  challenges.
\newblock {\em IEEE Signal Processing Magazine}, 31(1):34--44, 2013.

\bibitem{kolda2009tensor}
Tamara~G Kolda and Brett~W Bader.
\newblock Tensor decompositions and applications.
\newblock {\em SIAM review}, 51(3):455--500, 2009.

\bibitem{decurninge2020tensor}
Alexis Decurninge, Ingmar Land, and Maxime Guillaud.
\newblock Tensor-based modulation for unsourced massive random access.
\newblock {\em IEEE Wireless Communications Letters}, 2020.

\bibitem{sun2020opportunities}
Weike Sun and Richard~D Braatz.
\newblock Opportunities in tensorial data analytics for chemical and biological
  manufacturing processes.
\newblock {\em Computers \& Chemical Engineering}, page 107099, 2020.

\end{thebibliography}
